\documentclass[10pt,twocolumn,letterpaper]{article}

\usepackage{cvpr}

\usepackage[numbers]{natbib}

\definecolor{cvprblue}{rgb}{0.21,0.49,0.74}
\usepackage[breaklinks,colorlinks,allcolors=cvprblue]{hyperref}

\usepackage{multirow}
\usepackage{pifont}
\usepackage[accsupp]{axessibility}
\usepackage{booktabs}
\usepackage[ruled,vlined,linesnumbered]{algorithm2e}
\usepackage{algorithmic}
\usepackage{amsmath}
\usepackage{amssymb}
\usepackage{float}

\SetCommentSty{rmfamily}
\SetNlSty{rmfamily}{}{}

\usepackage{bibunits}
\defaultbibliographystyle{ieeenat_fullname}

\newcommand{\method}{\text{MotionEnhancer}}

\newcommand{\resetpaper}{
  \clearpage
  \setcounter{section}{0}
  \setcounter{subsection}{0}
  \setcounter{subsubsection}{0}
  \setcounter{figure}{0}
  \setcounter{table}{0}
  \setcounter{equation}{0}
  \setcounter{footnote}{0}
}

\makeatletter
\newcommand{\makesupplementarytitle}{
  \twocolumn[
    \begin{@twocolumnfalse}
    \centering
        {\Large \bfseries MotionEnhancer: Leveraging Video Diffusion for Motion-Enhanced Vision-Language Models \par}
        \vspace{0.6em}
        {\Large Supplementary Material \par}
        \vspace{1.2em}
    \end{@twocolumnfalse}
  ]
}
\makeatother

\begin{document}

\begin{bibunit}






\def\paperID{29583} 
\def\confName{CVPR}
\def\confYear{2026}

\title{MotionEnhancer: Leveraging Video Diffusion for Motion-Enhanced Vision-Language Models}

\author{
  Yifan Xu$^{1,2}$, Chao Zhang$^{2}$\thanks{Corresponding author}, Ruifei Ma$^{2}$, Fei Gao$^{2}$, Zhifei Yang$^{3}$, Jiaxing Qi$^{1}$, Zhipeng Chen$^{4}$ \\
  $^{1}$School of Computer Science and Engineering, Beihang University \\
  $^{2}$Beijing Digital Native Digital City Research Center \\
  $^{3}$School of Computer Science, Peking University \\
  $^{4}$School of Artificial Intelligence, Beijing University of Posts and Telecommunications \\
  {\tt\small yifan$\_$xu@buaa.edu.cn, ariczhang2009@gmail.com} \\
  \url{https://motion-enhancer.github.io/}
}



\maketitle
\begin{abstract}

The new era has witnessed a remarkable capability to extend Vision-Language Models (VLMs) for tackling tasks of video understanding.
While current VLMs excel at event- or story-level understanding, their ability to capture fine-grained motion details remains limited, primarily due to their focus on high-level static semantic structures and macro-event logic.
In contrast, Video Diffusion Models (VDMs) are adept at modeling dynamic motion patterns, benefiting from large-scale video data and the intrinsic requirement of temporal generation.
In this paper, we introduce \method{}, a novel approach that leverages motion priors distilled from a powerful video diffusion model as auxiliary supervision to enhance the motion understanding capability of a VLM via attention alignment.
\method{} comprises two simple parameter-free modules, Motion-sensitive Head Selection (MHS) and Motion-salient Text Token Identification (MTTI), to directly extract and optimize motion-related attentions from the VDM in a computation-only manner.
\method{} provides a scalable solution for motion understanding without additional training parameters, modifications to existing architectures, or tool calling.
Extensive experiments demonstrate that \method{} can achieve consistent improvements over state-of-the-art VLMs on two motion-level video understanding benchmarks, especially on motion-related metrics.

\end{abstract}    
\section{Introduction}

\begin{figure}[t]
    \centering
    \includegraphics[width=1.0\linewidth]{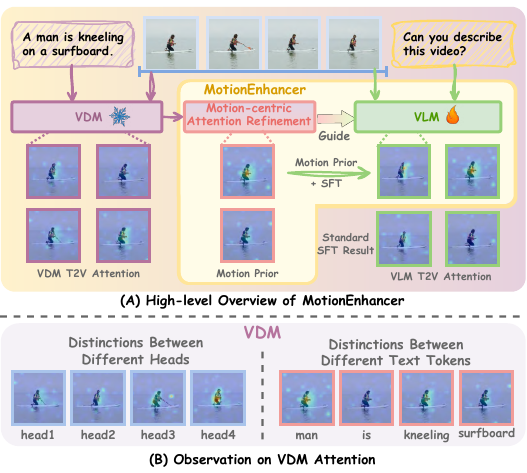}
    \caption{\textbf{(A) High-level overview of \method{}}, which incorporates motion priors from the VDM as guidance during supervised fine-tuning of the VLM for improved motion understanding.
    \textbf{(B) Observation of VDM attention.}
    We observe distinct patterns in the attention maps across different transformer heads and text tokens in the VDM, which motivates our refinement of motion-centric attention.
    }
    \label{fig:overview}
\end{figure}




In recent years, Vision-Language Models (VLMs) have become the mainstream framework for video understanding, advancing tasks like video captioning and question answering through multimodal alignment and semantic reasoning \cite{chen2024internvl2.5, zhang2025videollama3, li2024llavanext, comanici2025gemini, achiam2023gpt4, wang2024tarsier}.
Unlike static images, videos contain sequential frames that reflect scene dynamics over time.
The temporal relationships among these frames reveal how objects move, interact, and transform over time.
Effectively modeling these temporal relationships is crucial for capturing object movement and interaction.
Therefore, VLMs must understand not just individual frames, but also the dynamic changes across them.

Most video understanding VLMs follow a common pipeline: extract key frames, encode them with image encoders, and feed the features into multimodal models for alignment and reasoning \cite{bai2025qwen2, chen2024internvl2.5, zhang2025videollama3}.
As designed for understanding, the core competency of VLMs lies in capturing the overall meaning of a video by integrating information across frames, which prioritizes understanding high-level semantic structures such as static conceptual relationships and macro-event logic.
This task-driven approach enables strong performance on event- and story-level understanding, making VLMs well-suited for holistic tasks like video captioning.
However, they often overlook fine-grained motion details between frames, leading to a mismatch with the needs of motion-level understanding (for theoretical analysis, please see Sec. \ref{3.1theory}).

Meanwhile, Video Diffusion Models (VDMs) excel at generating visually realistic and temporally coherent videos \cite{blattmann2023stablevideodiffusion, yang2024cogvideox, wan2025wan, kong2024hunyuanvideo, zhang2025tora}.
During step-by-step denoising, VDMs learn the complex spatiotemporal patterns in videos, including physical laws of object motion, dynamic scene transitions, and inter-frame dependencies.
This inherent capability to capture real-world motion dynamics equips VDMs with a deep modeling of video motion patterns, making them serve as implicit simulators with learned interactive dynamics from large-scale video data.
Thus, video generative models unlock exciting possibilities for enhancing VLMs by providing more accurate motion modeling.
Since attention mechanisms \cite{vaswani2017attention} are central to VDMs, their text-to-vision attention naturally encodes motion priors.
Inspired by this, we propose \method{}, as shown in Fig. \ref{fig:overview}(A), which leverages motion priors distilled from a powerful VDM as auxiliary supervision to enhance the motion understanding capability of a VLM via simple attention alignment.
This idea aligns with advances in 2D vision-language tasks \cite{jin2025lavender}.
A key challenge arises:
\textit{How can we efficiently extract motion-focused attention signals from video QA pairs using a VDM?}

As shown in Fig. \ref{fig:overview}(B), attention maps from the VDM exhibit distinct patterns across transformer heads and text tokens.
To refine motion-centric attention, we propose two parameter-free modules: Motion-sensitive Head Selection (MHS) and Motion-salient Text Token Identification (MTTI).
MHS draws on the findings of SparseVideoGen \cite{xi2025sparsevideogen} and evaluates temporal attention maps using diagonal concentration, spatial continuity, and high-value region ratios to select motion-relevant heads.
MTTI computes frame-wise averages and inter-frame differences to identify text tokens responsive to both smooth and abrupt motion.
Notably, both modules require no extra training parameters.
They can directly extract and optimize motion-related attention maps from the VDM in a computation-only manner.

By aligning motion priors from the VDM with the text-to-vision attention in the VLM, our approach significantly improves the motion perception and reasoning ability of VLMs on motion-level benchmarks.
Crucially, it transfers capabilities from pretrained generative models rather than requiring access to their original training data, enabling an efficient form of capability transfer and enhancement that reduces dependence on large-scale video re-collection.
Our work demonstrates a new kind of cross-paradigm model interaction: using internal signals of one model family (e.g., generative VDMs) to guide another (e.g., discriminative VLMs).
Our focus is on introducing a simple and generalizable approach, rather than developing complex or specialized module designs for attention extraction or alignment.

Our main contributions can be summarized as follows:
\begin{itemize}
    \item We propose \method{}, a novel framework that leverages motion priors distilled from a powerful VDM as auxiliary supervision to enhance the motion understanding capability of a VLM through simple text-to-vision attention alignment.
    We also provide theoretical analysis to verify the feasibility of \method{}.
    \item To obtain effective motion priors, we design two parameter-free modules, MHS and MTTI, specifically adapted to the temporal characteristics of videos.
    They directly extract and optimize motion-related attention maps from the VDM in a computation-only manner.
    \item Extensive experiments on two motion-level video understanding benchmarks demonstrate the superiority of \method{}.
    Our results show that the attention alignment strategy can be successfully extended to video tasks with only minor adaptations.
\end{itemize}
\section{Related Work}

\begin{figure*}[t]
    \centering
    \includegraphics[width=1.0\linewidth]{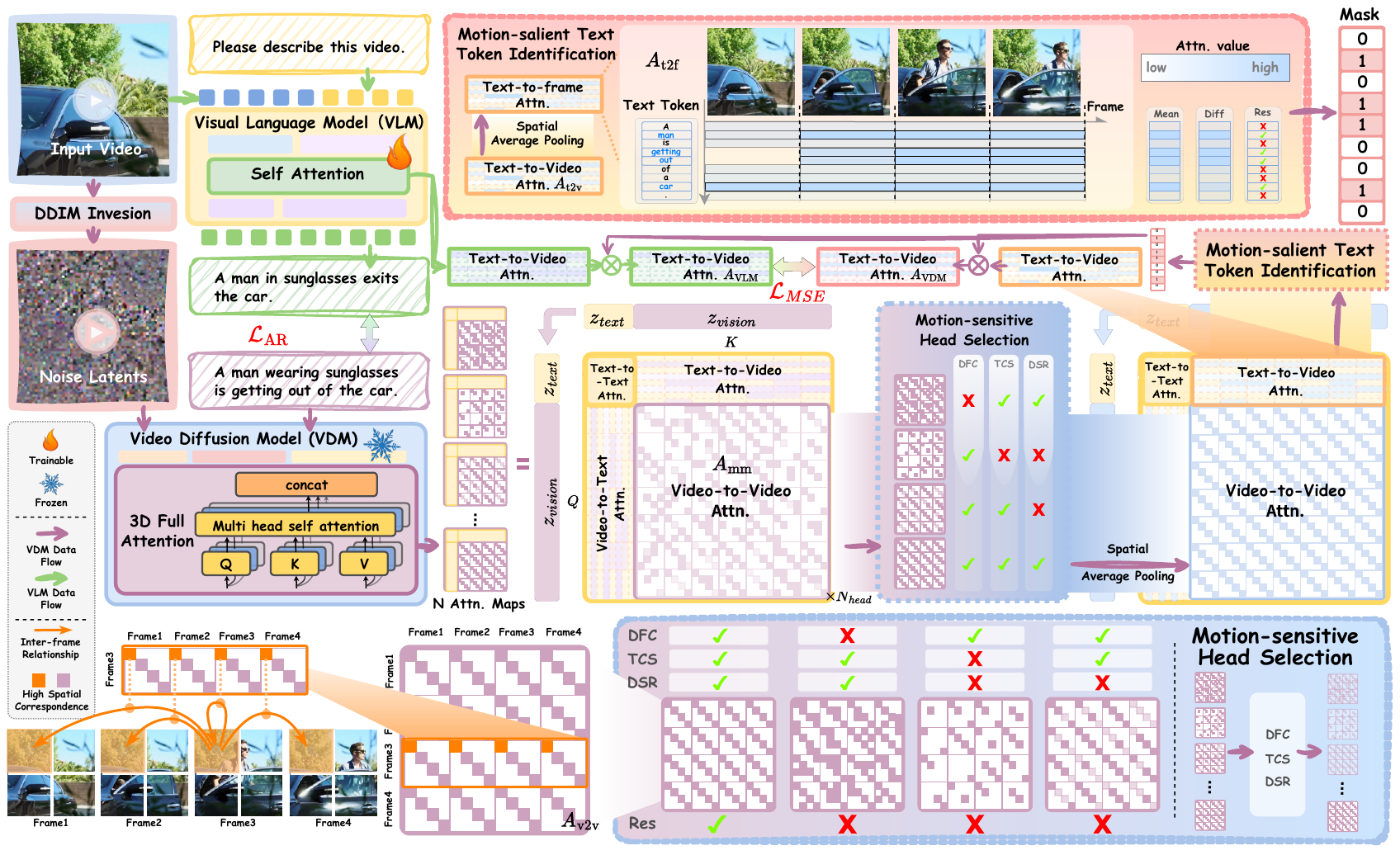}
    \caption{\textbf{Framework of \method{}}.
    Our method leverages motion priors distilled from a powerful VDM as auxiliary supervision to enhance the motion understanding capability of a VLM through attention alignment.
    Attention maps extracted from the VDM during DDIM sampling are filtered by the Motion-sensitive Head Selection (MHS) and Motion-salient Text Token Identification (MTTI) modules to identify motion-relevant attentions.
    The resulting text-to-vision attentions are then used to guide the VLM during supervised fine-tuning.}
    \label{fig:framework}
\end{figure*}

\subsection{VLMs for Video Understanding}

Recent advances in VLMs have prompted their adaptation to video understanding.
Most video VLMs include a visual encoder, modality alignment, and a Large Language Model (LLM) backbone.
A common strategy treats videos as image sequences by sampling key frames for encoding \cite{li2024llavanext, hong2024cogvlm2}.
Some enhance input flexibility through new positional embeddings and dynamic resolutions \cite{bai2025qwen2, chen2024internvl2.5}, while others use Q-Former \cite{li2023blip2}, spatial-temporal patchification \cite{liu2024kangaroo}, or adaptive pooling \cite{xu2024pllava} to compress and accelerate video encoding.
TE Fusion \cite{hong2025motionbench} groups frames and applies group-level self-attention for temporal modeling.
MotionSight \cite{du2025motionsight} improves motion modeling via object spotlighting and motion blur.

Despite progress, current VLMs remain limited in fine-grained motion understanding or rely on extra modules and external tools.
In contrast, we enhance VLM motion understanding by leveraging motion priors from a video diffusion model, without extra parameters or architecture changes.

\subsection{Diffusion Model Guidance for VLMs}

Diffusion models generate high-quality visual representations by progressively denoising latent features, capturing fine-grained semantics and structure \cite{rombach2022stablediffusion, blattmann2023stablevideodiffusion}.
This requires precise cross-modal modeling, allowing diffusion models to enhance VLMs' feature extraction and reasoning.
For instance, DIVA \cite{wang2024diva} uses generative feedback from frozen diffusion models to optimize CLIP features.
GenHancer \cite{ma2025genhancer} introduces lightweight denoisers and class-token-based reconstruction.
Lavender \cite{jin2025lavender} aligns VLM attention with Stable Diffusion to transfer visual expertise.

However, these methods focus mainly on images.
In this work, we explore a simple yet effective way to enhance VLM's ability for video motion understanding.
We show that attention map alignment between diffusion models and VLMs extends easily to video, requiring minimal temporal adaptation and no extra backbone changes.

\section{MotionEnhancer Theory}

We start by comparing VLM's learned distribution with the one required by motion understanding, and show that the latter naturally aligns with VDM's generative distribution.

\subsection{Distributional Mismatch Between VLMs and the Requirements of Motion Understanding}\label{3.1theory}

A VLM is trained with an autoregressive objective
\begin{equation}
    \mathcal{L}_{\text{AR}}=-\sum_i\log p_\theta(r_i\mid\mathbf{V},r_{<i}),
\end{equation}
where $\mathbf{V}=(V_1,\ldots,V_F)$ is the video and $r_i$ is the $i$-th response token.
Internally, its text-to-video attention can be interpreted as a discriminative conditional distribution
\begin{equation}
    A^{VLM}(t,s,f)\approx p_\theta(t\mid V_f(s),s_{<i}),
\end{equation}
where $t$ is a text token and $V_f(s)$ is the visual feature at spatial position $s$ in frame $f$.
This defines a $p(t|V)$ distribution, answering: \textit{“Given the visual input (often a single frame), how likely is this token?”}
Crucially, the model can satisfy this objective using static appearance cues, such as context and background, without modeling temporal dynamics or how visual evidence changes over time.

In contrast, motion understanding requires a different form of reasoning.
Motion-level questions $q$ like \textit{“Does the person start running then stop?”} or \textit{“In which direction does the camera move?”} depend on identifying where and when motion-related evidence appears across frames \cite{lei2020tvqa+, lu2025vited}.
This corresponds to a latent-variable factorization:
\begin{equation}
    p^{\star}(a\mid\mathbf{V},q)=\sum_\mathbf{E}p^{\star}(a\mid\mathbf{E},q)p^{\star}(\mathbf{E}\mid\mathbf{V},q),
\end{equation}
where $\mathbf{E}$ denotes motion-related latent evidence (object trajectories, motion segments, frame intervals) and $a$ is the answer.
To reveal the structure of $p^\star(\mathbf{E}|\mathbf{V},q)$, we follow common practices in grounded captioning and VideoQA \cite{mavroudi2022weakly, li2023discovering}, and decompose the global evidence $\mathbf{E}$ into per-token evidence subsets:
\begin{equation}
\mathbf{E}={E_t \mid t\in\mathcal{T}(q,a)},
\end{equation}
where $\mathcal{T}(q,a)$ is the set of semantic units, such as verbs and motion-related tokens, in the question or answer space.
Under this factorization, the evidence distribution becomes:
\begin{equation}
p^\star(\mathbf{E}\mid\mathbf{V},q)
=
\prod_{t\in\mathcal{T}(q,a)}
p^\star(E_t\mid\mathbf{V},t,q),
\end{equation}
reflecting the principle that the evidence relevant to each semantic unit $t$ is conditionally independent given the video.

Now, each $E_t$ must specify the spatiotemporal regions in $\mathbf{V}$ where the concept $t$ is visually grounded.
Assuming $E_t$ is selected by scoring each video location $(s,f)$ with a relevance function $r_t(s,f)$, we can rewrite:
\begin{equation}
p^\star(E_t\mid\mathbf{V},t,q)
\propto
\prod_{(s,f)\in E_t}
r_t(s,f),
\end{equation}
and the relevance score itself is a normalized conditional likelihood \cite{mavroudi2022weakly}:
\begin{equation}
r_t(s,f)
=
p^\star(v_{s,f}\mid t),
\end{equation}
which is the probability that the patch $v_{s,f}$ is an instance of the concept $t$.
Thus, $p^\star(\mathbf{E}|\mathbf{V},q)$ is governed by a set of concept-conditioned visual distributions $p^\star(v_{s,f}|t)$, where $t \in \mathcal{T}(q,a)$ and $(s,f) \in \mathbf{V}$.
This defines an evidence-seeking $p(V|t)$ distribution:
\textit{“Given a semantic concept $t$, where in the video is its visual evidence, and how does it evolve over time?”}

For semantic units that encode actions or dynamics, the associated latent evidence $\mathbf{E}$ must encode frame-to-frame transitions.
At the level of per-location features, this evidence can be expressed or alternated as motion:
\begin{equation}
    \mathbf{E}\textbf{[}\mathrm{Motion}(s,f)\textbf{]}\propto\|V_{f+1}(s)-V_f(s)\|.
\end{equation}

In summary, the distribution learned by VLMs ($p(t|V)$) is misaligned with the $p(V|t)$ structure required for motion understanding.
This mismatch explains VLMs’ strong appearance bias and limited temporal sensitivity, motivating the pursuit of the evidence-seeking distribution $p(V|t)$.

\subsection{Why VDM Attention is a Reliable Source of Motion Priors?}

A VDM naturally yields the concept-conditioned visual distribution for motion reasoning.
During generation, it predicts the clean latent video $\mathbf{z}_0$ from noisy inputs $\mathbf{z}_k$, conditioned on text tokens.
At each step, the model uses attention to determine how the visual content for a token $t$ should appear across spatial and temporal positions.
This attention guides what to generate at each location $(s,f)$, effectively forming a generative, concept-conditioned map \cite{tang2023daam}:
\begin{equation}
    A^{VDM}(t,s,f)\approx p_\phi(v_{s,f}\mid t,\mathbf{z}_k),
\end{equation}
reflecting where the semantics of token $t$ are visually realized in the generated video.

Unlike discriminative VLMs that can rely on static cues, VDMs must generate temporally coherent videos, ensuring adjacent frames form plausible motion.
When a token $t$ corresponds to dynamic content—like actions or camera movement—the denoising objective enforces accurate modeling of feature changes across frames.
Regions with higher temporal variation (larger $|V_{f+1}(s)-V_f(s)|$) are harder to reconstruct and thus receive more modeling focus.
As a result, the cross-attention for motion-related tokens becomes sensitive to temporal changes, highlighting regions whose dynamics align with the semantics of $t$.

This gives VDM attention two key properties.
First, it approximates the evidence-seeking distribution $p^\star(V|t)$ by highlighting where concept $t$ is grounded across space and time.
Second, it is naturally motion-calibrated, with attention shifts reflecting actual motion magnitudes.
Together, these properties make VDM-distilled attention effective for improving motion-enhanced VLMs via attention alignment.

\section{MotionEnhancer Methodology}

We propose leveraging motion priors from a VDM as auxiliary supervision to enhance VLMs’ motion understanding via attention alignment (Fig. \ref{fig:framework}).
Using DDIM inversion-sampling, we extract attention maps from the VDM and apply two parameter-free modules, Motion-sensitive Head Selection (MHS) and Motion-salient Text Token Identification (MTTI), to identify motion-relevant heads and tokens.
These motion priors guide attention alignment in the VLM.

While we use Qwen2.5-VL \cite{bai2025qwen2}, InternVL3 \cite{zhu2025internvl3}, and CogVideoX-1.5-5B \cite{yang2024cogvideox}, our method is general and applicable to other VLMs and DiT-based VDMs.

\subsection{Video Diffusion-based Attention Extraction}

An input video is compressed by a 3D causal VAE into a latent $z_{\text{vision}}\in \mathrm{R}^{F \times H \times W}$, where $F$, $H$, and $W$ denote frames, height, and width, with total sequence length $S = F \times H \times W$.
We apply 5-step DDIM inversion (details in supplementary) to extract noise, which is concatenated with text embeddings $z_{\text{text}}$ to form the multimodal sequence $z_{\text{mm}}$.
CogVideoX is trained with zero terminal SNR, following LDM’s noise schedule \cite{rombach2022stablediffusion}, improving generation quality but still allowing some sampling deviations \cite{yesiltepe2025dynamic}.
To address this, we reconstruct from noise using classifier-free guidance, incorporating cross-stream memory from a parallel DDIM-inverted path.
During the 5-step denoising, attention maps \cite{vaswani2017attention} are computed and stored at each step:
\begin{equation}
    A_{\text{mm}} = \text{Softmax} \left(\frac{Q_{\text{mm}}K_{\text{mm}}^T}{\sqrt{d}}\right),
\end{equation}
where $Q_{\text{mm}}$ and $K_{\text{mm}}$ are the queries and keys from $z_{\text{mm}}$, and $d$ is the feature dimension.
The final attention map $A_{\text{mm}}$ is obtained by applying both layer-wise and timestep-wise average pooling to the raw attention outputs.

\subsection{Motion-centric Attention Refinement}

For each transformer head in the VDM, we extract its attention map $A_{\text{mm}}$.
Since not all heads capture temporal motion \cite{xi2025sparsevideogen}, we propose Motion-sensitive Head Selection (MHS) to select and aggregate motion-relevant heads.
Additionally, Motion-salient Text Token Identification (MTTI) filters out motion-irrelevant text, helping the VLM focus on meaningful text-video motion connections.

\noindent \textbf{Motion-sensitive Head Selection (MHS).}
To identify motion-relevant heads, we leverage the observation that attention weights often form diagonal patterns in frame-level maps, indicating temporal continuity in fixed regions \cite{xi2025sparsevideogen, ma2025followyourmotion}.
We quantify this pattern without introducing trainable parameters.
For each head, a motion mask $\mathcal{M} \in \mathrm{R}^{S \times S}$ captures this diagonal structure.
We then evaluate each vision-to-vision attention map $A_{\text{v2v}} \in \mathrm{R}^{S \times S}$ using three metrics: concentration, spatial coherence, and prevalence.

(1) Diagonal Focus Coefficient (DFC).
DFC measures the proportion of attention concentrated within the diagonal mask compared to regions outside the mask.
For the attention map $A_{\text{v2v}}$, DFC is defined as:
\begin{equation}
    \mathrm{DFC}=\frac{\sum_{(i,j)\in\mathcal{M}} A^2_{\text{v2v}}[i,j]}{\sum_{(i,j)\notin\mathcal{M}}A^2_{\text{v2v}}[i,j]}.
\end{equation}
A higher DFC indicates that more attention is focused along the diagonal, suggesting stronger motion relevance.

(2) Temporal Continuity Score (TCS).
TCS measures the persistence of high diagonal attention, indicating consistent spatial focus across frames.
We set the threshold $\tau$ as the average attention value.
For each spatial location $s$, we extract a cross-frame submatrix $A_s \in \mathrm{R}^{F \times F}$ from $A_{\text{v2v}}$, capturing attention between $s$ across all frame pairs.
In each row $f$ of $A_s$, we identify maximal contiguous segments where attention exceeds $\tau$.
Let $\text{Len}(s) = {l_1, l_2, ..., l_m}$ be these segment lengths.
TCS is the mean segment length across all $s$:
\begin{equation}
    \mathrm{TCS}=\frac{1}{S}\sum_{s=1}^S \frac{1}{m}\sum_{i=1}^ml_i,
\end{equation}
where $S$ is the total number of spatial locations, and if $\text{Len}(s)$ is empty, the segment length is set to 0. A higher TCS indicates more sustained attention to consistent regions across frames, reflecting stronger motion continuity.

(3) Diagonal Saliency Ratio (DSR).
DSR quantifies how frequently high attention appears in the diagonal region $D$.
Using the same threshold $\tau$ as in TCS, we count the number of entries $n_{\text{high}}$ in $D$ where attention $\geq \tau$.
Let $|D|$ be the total number of entries in $D$. Then, DSR is defined as:
\begin{equation}
    \mathrm{DSR}=\frac{n_{\mathrm{high}}}{|D|}.
\end{equation}
Larger DSR means high attention is more widespread along the diagonal, not just at isolated spots.

After computing DFC, TCS, and DSR for all heads, we standardize each metric using its mean and standard deviation.
Each head is assigned a composite score by summing its normalized metrics, and the top 50\% are selected as motion-related heads. We then average-pool the attention maps $A_{\text{mm}}$ from these selected heads.

\noindent \textbf{Motion-salient Text Token Identification (MTTI).}
After aggregating motion-aware heads, we extract the text-to-vision attention region, yielding $A_{\text{t2v}} \in \mathrm{R}^{T \times S}$.
Since not all text tokens relate to motion, we assess each token’s temporal dynamics.
By average pooling over spatial dimensions ($H \times W$), we obtain $A_{\text{t2f}} \in \mathrm{R}^{T \times F}$, where each row captures a token’s attention across frames.
For each token $t$, we compute the mean of its attention and the mean of its first-order differences. The motion score is defined as:
\begin{equation}
\footnotesize
    \text{MS}(t)={\mathrm{Mean}_f(A_{\text{t2f}}^t)} + \frac{1}{F-1}\sum_{f=1}^{F-1}|A_{\text{t2f}}^t(f+1)-A_{\text{t2f}}^t(f)|.
\end{equation}
The mean attention value captures a token’s overall importance, while the mean first-order difference reflects its temporal fluctuation—higher for dynamic events, lower for static elements.
We rank tokens by motion scores and select the top 50\% for alignment, ensuring they are both salient and temporally dynamic.
The resulting VDM attention is $A_{\text{VDM}} \in \mathrm{R}^{T' \times S}$, where $T'$ is the number of selected tokens.

\begin{table*}
\setlength{\tabcolsep}{0.9mm}
\centering
\caption{Quantitative results of MotionBench. * denotes results we reproduced using their open-source code, while other results are taken from the original benchmark.}
\begin{tabular}{lc|cc|cccccc} 
\toprule
Model                                                   & Frames & Overall                     & Average                     & MR    & LM    & CM    & MO    & AO    & RC     \\ 
\hline
\multicolumn{10}{l}{Small Size Series}                                                                                                                                                \\ 
\hline
Qwen2.5-VL-3B* \cite{bai2025qwen2}                                         & 1fps   & 53.56                       & 49.45                       & 59.54 & 53.11 & 38.44 & 70.14 & 40.46 & 35.00  \\
\textbf{Qwen2.5-VL-3B + MotionEnhancer (Ours)}          & 1fps   & $\text{56.60}^{\uparrow3.04}$ & $\text{52.51}^{\uparrow3.06}$ & 63.06 & 61.72 & 47.01 & 68.84 & 43.16 & 31.25  \\
InternVL3-2B* \cite{zhu2025internvl3}                                          & 8      &     53.96                        &    49.69                         &   60.01    &  57.69     &  43.90     &   70.00    &   40.27    &   26.25     \\
\textbf{InternVL3-2B + MotionEnhancer (Ours)} & 8      &            $\text{55.50}^{\uparrow1.54}$                 &          $\text{51.35}^{\uparrow1.66}$                  &  61.57     &   57.51    &   46.23    &   71.30    &  42.00     &   29.50     \\ 
\hline
\multicolumn{10}{l}{Medium Size Series}                                                                                                                                                \\ 
\hline
MiniCPM-V2.6-7B \cite{yao2024minicpm}                                        & 64     & 52                          & -                           & -     & -     & -     & -     & -     & -      \\
GLM4-9B + TE Fusion \cite{hong2025motionbench}                                    & 16     & 58                          & -                           & -     & -     & -     & -     & -     & -      \\
Qwen2.5-VL-7B* \cite{bai2025qwen2}                                         & 1fps   & 52.81                       & 48.29                       & 59.00 & 54.58 & 35.58 & 71.30 & 38.54 & 30.75  \\
Qwen2.5-VL-7B + MotionSight* \cite{du2025motionsight}                           & 1fps   & 55.30                       & 51.56                       & 59.88 & 57.33 & 47.01 & 73.91 & 40.46 & 30.75  \\
\textbf{Qwen2.5-VL-7B + MotionEnhancer (Ours)}          & 1fps   & $\text{57.04}^{\uparrow4.23}$ & $\text{52.92}^{\uparrow4.63}$ & 63.40 & 61.54 & 47.27 & 70.29 & 43.55 & 31.50  \\
InternVL3-8B* \cite{zhu2025internvl3}                                          & 8      & 54.88                       & 50.81                       & 60.42 & 58.06 & 43.64 & 70.29 & 43.93 & 28.50  \\
\textbf{InternVL3-8B +MotionEnhancer (Ours)} \cite{zhu2025internvl3}                                          & 8      & $\text{57.69}^{\uparrow2.81}$                       & $\text{53.22}^{\uparrow2.41}$                       & 64.14 & 60.07 & 48.83 & 75.94 & 40.85 & 29.50  \\
\hline
\multicolumn{10}{l}{Large Size Series}                                                                                                                                            \\ 
\hline
PLLaVA-34B \cite{xu2024pllava}                                             & 16     & 52                          & -                           & -     & -     & -     & -     & -     & -      \\
Qwen2.5-VL-72B* \cite{bai2025qwen2}                                        & 1fps   & 58.30                       & 54.32                       & 64.00 & 60.30 & 48.60 & 73.20 & 46.80 & 33.00  \\
\bottomrule
\end{tabular}
\label{Tab:MotionBench}
\end{table*}

\subsection{Attention Alignment}

The VLMs employ self-attention to model semantic and spatial relationships between text and visual tokens across multiple heads and layers.
Following the same procedure as with the VDM, we apply average pooling across heads and layers to obtain an attention matrix $A_{\text{VLM}} \in \mathrm{R}^{T' \times S}$, where each row reflects how a text token attends to visual patches.

At this point, we have the text-to-vision attention maps from both the VDM and VLM, denoted as $A_{\text{VDM}}$ and $A_{\text{VLM}}$.
We first interpolate $A_{\text{VLM}}$ to match the dimensions of $A_{\text{VDM}}$, then use a 3-layer MLP as the aligner network.
\begin{equation}
    \mathcal{L}_{\text{MSE}} = ||\text{Aligner}(A_{\text{VLM}})-A_{\text{VDM}}||_2,
\end{equation}
where $||\cdot||_2$ denotes L2-norm.
Note that only the previously selected text tokens are involved in this alignment step.

The total loss is as below and optimized in a supervised fine-tuning (SFT) manner, with $\lambda$ as a balance factor:
\begin{equation}
    \mathcal{L}_{\text{total}} = \mathcal{L}_{\text{AR}} + \lambda\mathcal{L}_{\text{MSE}}.
\end{equation}

\subsection{Discussion}\label{Sec:discussion}
Here, we discuss two questions:
(1) \textit{Why do we use average pooling for VLM heads instead of motion-based head selection?}
VDMs are generative and their heads specialize in spatial or temporal aspects. In contrast, VLMs focus on understanding and lack such clear specialization.
Thus, average pooling is more suitable for VLMs.
As a result, VLM transformer heads are more general-purpose and do not show the clear specialization found in VDMs.

(2) \textit{Can the text token identification strategy select only motion-related tokens?}
While aimed at selecting motion-relevant tokens, motion is often carried by verbs and their associated subjects or objects.
Our method mainly filters out unrelated tokens (e.g., function words like "the", "which") rather than isolating only verbs.


\section{Experiments}

\subsection{Experimental Setups}

\textbf{Training Data.}
We leverage all 5k video QA pairs from MotionBench-Train \cite{hong2025motionbench} and sample 20k pairs from MotionVid-QA \cite{du2025motionsight}, totally forming 25k pairs for training.

\noindent \textbf{Motion-level Benchmarks.}
We evaluate our approach on two motion-level video understanding benchmarks: MotionBench \cite{hong2025motionbench} and FAVOR-Bench \cite{tu2025favorbench}.

MotionBench includes 5,385 videos and 8,052 QA pairs on six motion-focused tasks: Motion Recognition (MR), Location-related Motion (LM), Action Order (AO), Repetition Count (RC), Motion-related Objects (MO), and Camera Motion (CM).
We use its official Dev set for evaluation.

FAVOR-Bench’s close-ended test set contains 1,776 videos and 8,184 QA pairs spanning six dimensions: Action Sequence (AS), Holistic Action Classification (HAC), Single Action Detail (SAD), Multiple Action Details (MAD), Camera Motion (CM), and Non-Subject Motion (NSM).

\noindent \textbf{Evaluation metrics.}
We report results as accuracy scores for each task type.
`Overall' accuracy is computed across all questions, reflecting total performance, while the `Average' metric refers to the mean accuracy over all types, ensuring equal consideration for each type of motion problem regardless of how many samples each contains.

\begin{table*}
\setlength{\tabcolsep}{0.9mm}
\centering
\caption{Quantitative results of FAVOR-Bench. * denotes results we reproduced using their open-source code, while other results are taken from the original benchmark. (For more VLMs, please see supplementary materials.)}
\begin{tabular}{lc|cc|cccccc} 
\toprule
Model                         & Frames & Overall & Average & AS    & HAC   & SAD   & MAD    & CM     & NSM    \\ 
\hline
\multicolumn{10}{l}{Small Size Series}                                                                                \\ 
\hline
VideoLLaMA3-2B \cite{zhang2025videollama3}               & 1fps   & 32.98~  & 34.61   & 28.97 & 36.60 & 34.90 & 38.01  & 28.56  & 40.62  \\
Qwen2.5VL-3B* \cite{bai2025qwen2}                 & 1fps   & 37.43~  & 38.07   & 38.45 & 38.16 & 39.35 & 43.40  & 23.72  & 45.31  \\
\textbf{Qwen2.5VL-3B + MotionEnhancer (Ours)} & 1fps   & $\text{44.53}^{\uparrow7.10}$   & $\text{43.94}^{\uparrow5.87}$   & 45.01 & 51.59 & 44.40 & 48.96  & 28.37  & 45.31  \\ 
InternVL3-2B* \cite{zhu2025internvl3}                 & 8   & 39.27~  & 39.11   & 37.66 & 43.28 & 40.49 & 44.98  & 29.21  & 39.06  \\
\textbf{InternVL3-2B + MotionEnhancer (Ours)} & 8   & $\text{43.71}^{\uparrow4.44}$   & $\text{45.35}^{\uparrow6.24}$   & 38.53 & 54.57 & 42.60 & 51.78  & 33.02  & 51.56  \\ 
\hline
\multicolumn{10}{l}{Medium Size Series}                                                                                \\ 
\hline
LLaVA-Video-7B-Qwen2 \cite{zhang2024llavavideo}         & 64     & 38.60   & 39.94   & 36.14 & 41.27 & 41.28 & 44.48  & 29.58  & 46.88  \\
VideoChat-Flash-Qwen2-7B \cite{li2024videochat}     & 1fps   & 43.82   & 44.86   & 41.90 & 48.41 & 42.84 & 50.95  & 35.07  & 50.00  \\
VideoLLaMA3-7B \cite{zhang2025videollama3}               & 1fps   & 41.46   & 41.46   & 40.20 & 44.13 & 42.42 & 48.30  & 31.53  & 42.19  \\
Qwen2.5VL-7B* \cite{bai2025qwen2}                 & 1fps   & 42.61   & 42.58   & 41.64 & 47.83 & 44.89 & 47.55  & 28.28  & 45.31  \\
Qwen2.5VL-7B + MotionSight* \cite{du2025motionsight}  & 1fps   & 45.47   & 45.99   & 46.23 & 51.59 & 45.01 & 50.04  & 29.95  & 53.12  \\
\textbf{Qwen2.5VL-7B + MotionEnhancer (Ours)} & 1fps   & $\text{46.88}^{\uparrow4.27}$   & $\text{47.01}^{\uparrow4.43}$   & 49.34 & 50.62 & 45.37 & 53.20  & 30.42  & 53.12  \\ 
InternVL3-8B* \cite{zhu2025internvl3}                 & 8   & 45.82   & 46.35   & 45.39 & 48.54 & 47.59 & 51.45  & 33.58  & 51.56  \\
\textbf{InternVL3-8B + MotionEnhancer (Ours)} & 8   & $\text{48.94}^{\uparrow3.12}$   & $\text{49.25}^{\uparrow2.90}$   & 47.17 & 57.11 & 46.57 & 56.35  & 36.74  & 51.56  \\ 
\hline
\multicolumn{10}{l}{Large Size Series}                                                                            \\ 
\hline
LLaVA-Video-72B-Qwen2 \cite{zhang2024llavavideo}        & 64     & 46.08   & 46.49   & 48.35 & 47.50 & 45.25 & 51.70  & 33.02  & 53.12  \\
Qwen2.5-VL-72B* \cite{bai2025qwen2}               & 1fps   & 48.14   & 48.17   & 50.28 & 46.98 & 48.13 & 51.78  & 40.28  & 51.56  \\
\bottomrule
\end{tabular}
\label{Tab:FAVOR-Bench}
\end{table*}

\subsection{Implementation Details}

We use CogVideoX-1.5-5B \cite{yang2024cogvideox} as the VDM, and Qwen2.5-VL (3B, 7B) \cite{bai2025qwen2} and InternVL3 (2B, 8B) \cite{zhu2025internvl3} as VLMs.
Attention maps are extracted from the frozen VDM via 5-step DDIM sampling after 5-step inversion.
This process is fully offline before VLM SFT, and the priors can be reused across VLMs and ablations.
In practice, one-time extraction takes 20-30 seconds on an A100 GPU.
During SFT, the vision tower, merger, and LLM backbone are trainable.
We use AdamW \cite{loshchilov2017adamw} with $\beta_1=0.9$, $\beta_2=0.999$, $\epsilon=1\mathrm{e}{-8}$, and weight decay 0.1, and a cosine scheduler with 0.03 warmup ratio.
Learning rates are $1\mathrm{e}{-5}$ for the LLM and merger, and $2\mathrm{e}{-6}$ for the vision tower.
The loss factor $\lambda$ is set to 1.
Training runs for one epoch with batch size 8 on eight A100 GPUs (80GB) using DeepSpeed.

\begin{table}
\centering
\caption{Ablation study of MHS and MTTI using Qwen2.5VL-7B.
These results confirm that MHS and MTTI are complementary, and combining them yields the highest gains.}
\begin{tabular}{c|cc|cc|cc} 
\toprule
\multirow{2}{*}{Idx} & \multicolumn{2}{c|}{Variants} & \multicolumn{2}{c|}{MotionBench} & \multicolumn{2}{c}{FAVOR-Bench}  \\ 
\cline{2-7}
                       & MHS & MTTI                    & Over. & Aver.                & Over. & Aver.                \\ 
\hline
1                      & \ding{55} & \ding{55}                     & 54.83     & 51.51                    & 44.83     & 44.54                    \\
2                      & \ding{51} & \ding{55}                     & 56.60     & 52.51                    & 46.65     & 46.55                    \\
3                      & \ding{55} & \ding{51}                     & 55.80     & 51.31                    & 45.47     & 45.99                    \\
4                      & \ding{51} & \ding{51}                     & 57.04      & 52.92                    & 46.88       & 47.01                    \\
\bottomrule
\end{tabular}
\label{Tab:Ablation}
\end{table}

\subsection{Comparison with State-of-the-art Method}

\textbf{Baselines.}
Our compared methods are categorized based on model size: small, medium and large size series, including popular open-source VLMs and improvements on them.
We perform comparison in small and medium series, while showing the larger series as the upper limit of this task.

\begin{figure*}
    \centering
    \includegraphics[width=1.0\linewidth]{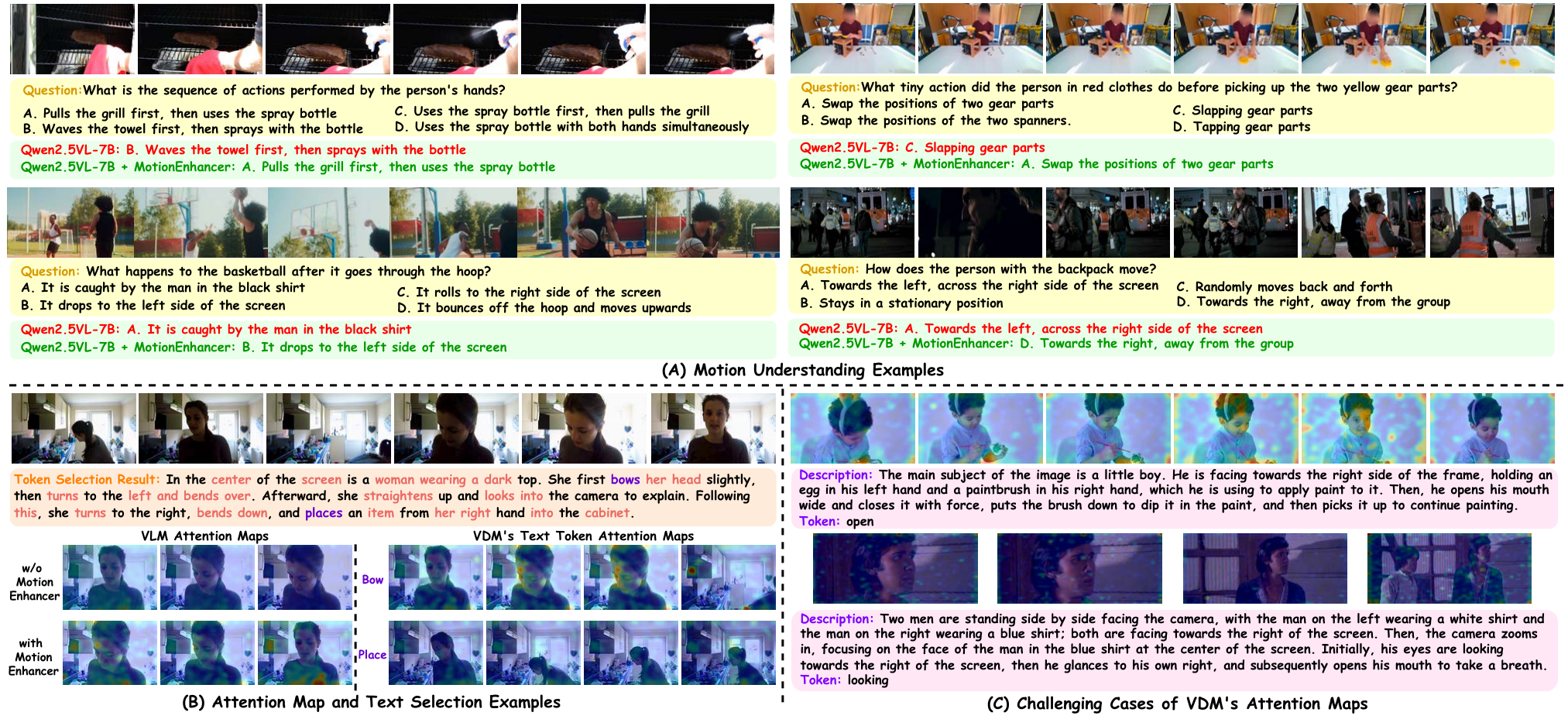}
    \caption{Qualitative examples of \method{}. (More examples can be found in supplementary materials.)}
    \label{fig:cases}
\end{figure*}

\noindent \textbf{MotionBench.}
Tab. \ref{Tab:MotionBench} presents the evaluation results on MotionBench.
Incorporating MotionEnhancer consistently yields substantial improvements for Qwen2.5-VL across both 3B and 7B variants.
For the 3B backbone, MotionEnhancer elevates the category average by 3.1\%, with pronounced gains in motion-relevant metrics.
For the 7B backbone, the enhancement effect becomes more significant, with category average increasing by 4.6\%, and MR and CM improving by 4.4\% and 11.7\%, respectively.
We observe similar improvements with InternVL3 backbone, with an overall accuracy of 1.51\% for the 2B backbone and 2.81\% for the 8B backbone.
These results indicate that MotionEnhancer effectively augments temporal motion modeling.



\noindent \textbf{FAVOR-Bench.}
Tab. \ref{Tab:FAVOR-Bench} reports the results on FAVOR-Bench, further validating the effectiveness of MotionEnhancer for fine-grained motion understanding.
Across backbone sizes, Qwen2.5-VL-3B and 7B derive consistent and substantial gains from MotionEnhancer, and surpass other models in their respective series.
On the 3B backbone, MotionEnhancer improves category average by 5.6\% and overall score by 7.1\%, with remarkable gains in HAC and MAD.
On the 7B backbone, category average rises by 4.5\% and overall score by 4.2\%, with consistent gains in AS, HAC, and MAD.
We observe similar improvements with InternVL3 backbone, with an overall accuracy of 4.44\% for the 2B backbone and 3.12\% for the 8B backbone.

Notably, Qwen2.5-VL-3B+MotionEnhancer surpasses Qwen2.5-VL-7B on both benchmarks, while Qwen2.5-VL-7B+MotionEnhancer achieves performance comparable to Qwen2.5-VL-72B and other large-scale baselines.
These results highlight that MotionEnhancer scales robustly across backbone sizes, enabling compact models to attain performance levels typically associated with substantially larger architectures.


\subsection{Ablation Study}

Table \ref{Tab:Ablation} presents the ablation results of MHS and MTTI on Qwen2.5VL-7B.
The baseline (Index 1) includes neither module, and is trained only on our 25k dataset, using average pooling for both motion heads and text tokens during alignment.
It achieves Overall/Average scores of 54.83/51.51 on MotionBench and 44.83/44.54 on FAVOR-Bench.
Adding only MHS (Index 2) brings significant improvements: +1.77 Overall on MotionBench and +1.82 Overall on FAVOR-Bench, showing that selecting motion-relevant heads effectively enhances temporal motion modeling.
Applying only MTTI (Index 3) also consistently improves over the baseline, indicating that aligning with motion-related text tokens benefits motion understanding.
However, its gain is smaller than that of MHS, since MTTI benefits from the prior motion filtering provided by MHS.
Combining both MHS and MTTI (Index 4) achieves the best performance, with Overall/Average scores of 57.04/52.92 on MotionBench and 46.88/47.01 on FAVOR-Bench.
These results confirm that MHS and MTTI are complementary, and together yield the largest gains.

\subsection{Qualitative Examples and Limitations}

\textbf{Qualitative Examples.}
We provide qualitative examples of motion understanding, attention, and challenging cases.
In Fig.~\ref{fig:cases}(A), MotionEnhancer helps better answer motion-related questions.
Fig.~\ref{fig:cases}(B) shows that MHS improves motion-focused attention, while MTTI effectively filters out irrelevant tokens (see Sec.~\ref{Sec:discussion}).
After alignment, the VLM attends more to motion cues and related objects.

\noindent \textbf{Limitations and future improvements.}
In practice, after training, we observe fewer wrong-to-correct cases for videos in which the main subject fills the frame and remains static.
To understand the reason, we visualize the VDM attention maps of these challenging videos in Fig. \ref{fig:cases}(C), and find that the attention becomes diffuse and less focused, failing to highlight specific objects or motions.
This limitation stems from a bias inherent in the VDM training data.
Since the VDM is mainly trained on videos containing small objects, it struggles to model large, static subjects that occupy the entire frame.
Future work could explore more refined motion extraction methods and introduce data preprocessing strategies to mitigate this bias.

\vspace{-2mm}
\section{Conclusion}

In this work, we introduce \method{}, which leverages motion priors distilled from a powerful VDM as auxiliary supervision to enhance VLM's motion understanding capability via attention alignment.
Extensive experiments show that \method{} consistently improves over state-of-the-art VLMs on two motion-level video understanding benchmarks, especially on motion-related metrics, demonstrating \method{} provides a scalable solution for motion understanding without requiring extra training parameters and architectural modifications.
Moreover, motion latents extracted from large-scale videos via VDMs can serve as a motion-aware pretraining signal for downstream tasks that are highly sensitive to temporal dynamics (e.g., robotic arm grasping), offering strong VLM initialization that improves sample efficiency and temporal generalization.
We leave this for future exploration.




{\small
\putbib[main]
}

\end{bibunit}

\resetpaper

\begin{bibunit}






\def\paperID{*****} 
\def\confName{CVPR}
\def\confYear{2026}





\clearpage
\setcounter{page}{1}
\makesupplementarytitle
\section{Preliminaries}

\subsection{DDIM Framework}

Denoising Diffusion Implicit Models (DDIM) \cite{song2020ddim} provide an efficient and flexible sampling framework for diffusion models.
Traditional diffusion models, such as Denoising Diffusion Probabilistic Models (DDPM) \cite{ho2020ddpm}, rely on a stochastic generation process, where random noise is gradually removed through a series of probabilistic denoising steps.
This approach, while effective, often requires a large number of iterations to generate high-quality samples, resulting in high computational cost.
DDIM, in contrast, introduces a deterministic alternative to the sampling process. It leverages a non-Markovian forward process that adds Gaussian noise to the original data $\mathbf{x}_0$ according to a predefined noise schedule ${\alpha_t}$.
Specifically, the forward process can be formulated as:
\begin{equation}
\mathbf{x}_t = \sqrt{\alpha_t}\mathbf{x}_0 + \sqrt{1 - \alpha_t}\epsilon, \quad \epsilon \sim \mathcal{N}(0, \mathbf{I}),
\end{equation}
where $\alpha_t$ decreases monotonically over time, gradually corrupting the data.

The key innovation of DDIM lies in its reverse process. Instead of sampling new noise at every step, DDIM uses a deterministic mapping to reconstruct the data. The denoising step is given by:
\begin{equation}
\mathbf{x}_{t-1} = \sqrt{\alpha_{t-1}}\hat{\mathbf{x}}_0(\mathbf{x}_t) + \sqrt{1 - \alpha_{t-1}} \epsilon_\theta(\mathbf{x}_t, t),
\end{equation}
where $\hat{\mathbf{x}}_0(\mathbf{x}_t)$ is the predicted clean data:
\begin{equation}
\hat{\mathbf{x}}_0(\mathbf{x}_t) = \frac{\mathbf{x}_t - \sqrt{1 - \alpha_t}\epsilon_\theta(\mathbf{x}_t, t)}{\sqrt{\alpha_t}}.
\end{equation}

By removing the stochasticity from the reverse process, DDIM enables faster and more controllable sampling.
This deterministic formulation allows the model to generate high-quality samples with significantly fewer steps compared to DDPM, while preserving the overall data structure and ensuring boundary consistency.
As a result, DDIM has emerged as a widely favored option in diffusion-based generative modeling, especially within applications where both efficiency and fidelity are critical requirements.

\subsection{DDIM Inversion}
DDIM inversion \cite{mokady2023ddiminversion} extends the DDIM framework by enabling the mapping of real data $\mathbf{x}_0$ back into its corresponding latent noise $\mathbf{x}_T$.
This process is achieved through an iterative application of the reversed DDIM denoising step, effectively tracing the generative trajectory in the opposite direction.

Formally, starting from the observed data $\mathbf{x}0$, the inversion procedure sequentially estimates the intermediate noisy representations $\mathbf{x}_t$ for $t = 0, \ldots, T-1$ via:
\begin{equation}
    \mathbf{x}_{t+1} = \sqrt{\alpha_{t+1}}\hat{\mathbf{x}}_0(\mathbf{x}_t) + \sqrt{1-\alpha_{t+1}}\epsilon_\theta(\mathbf{x}_t,t).
\end{equation}
After $T$ iterations, this process yields a latent code $\mathbf{x}_T=\mathcal{G}^{-1}(\mathbf{x}_0)$, which encodes the original data within the noise space of the diffusion model.

A notable feature of DDIM inversion is its path determinism: if the noise prediction function $\epsilon_\theta$ remains fixed, the entire sequence is reversible.
In other words, if one starts from $\mathbf{x}_0$, performs the inversion step, and then follows with the standard forward generative process, the original input will be reconstructed exactly:
\begin{equation}
    \mathbf{x}_0 \xrightarrow{\mathrm{invert}} \mathbf{x}_T \xrightarrow{\mathrm{reconstruct}} \mathbf{x}_0.
\end{equation}
This deterministic and bijective mapping ensures that each data point has a unique latent representation and that the model’s internal states can be precisely analyzed.

Such inversion capability is especially valuable for investigating the internal mechanisms of diffusion models, enabling controllable editing, semantic manipulation, and deeper interpretability.
It provides a systematic approach to relate observable samples with their latent origins, thus opening avenues for fine-grained analysis and intervention in generative modeling tasks.

\subsection{Introduction of Our Used Models}

We utilize CogVideoX-5B as our Video Diffusion Model (VDM) and Qwen2.5-VL (3B and 7B) and InternVL3 (2B and 8B) as our Vision-Language Models (VLM).
Below is a concise overview of these models, covering their architectural designs and operational workflows.

\textbf{CogVideoX: A DiT-based Video Generative Model.}
CogVideoX \cite{yang2024cogvideox} is a diffusion-transformer model \cite{peebles2023dit} for text-to-video generation that tackles the twin challenges of high-resolution fidelity and long-range temporal coherence in a single, end-to-end pipeline.
Its first stage is a 3D causal VAE that compresses both spatial and temporal axes with an 8×8×4 reduction factor, turning seconds of raw pixels into a compact latent cube while preserving fine detail and motion continuity; the causal design guarantees that decoding can proceed frame-by-frame for streaming applications.
The latent volume is patched into spatio-temporal tokens and simply concatenated with T5 text embeddings \cite{raffel2020t5}, eliminating cumbersome cross-attention modules and allowing the same transformer blocks to process both modalities.
Inside each transformer layer, 3D full attention lets every patch attend to any other patch in its causal window, capturing large motions without drift. Separate expert adaptive LayerNorm branches scale and shift visual and textual features before they merge, aligning modalities without extra parameters.

\textbf{Qwen2.5-VL: Flagship VLM in Qwen Series.}
Qwen2.5-VL \cite{bai2025qwen2} is a VLM that ingests images and text at their native resolution without ever cropping them into fixed grids.
At its core is a redesigned Vision Transformer (ViT) \cite{dosovitskiy2020vit} trained from scratch, which incorporates window attention in most layers—with only four layers using full attention—to reduce computational complexity from quadratic to linear relative to input patches, enabling native handling of dynamic resolutions without normalization artifacts.
The resulting dense visual tokens are streamed into a lightweight vision-language merger that groups neighboring patches and projects the ensemble through a two-layer MLP, yielding a compact set of vision embeddings that align seamlessly with the text token space while retaining fine-grained spatial cues.
For video, Qwen2.5-VL introduces Multimodal Rotary Position Embedding (MRoPE) tied to absolute timestamps: temporal position IDs are explicitly synchronized with real-world clock time, so the model perceives frame-rate variations naturally and reasons about event order without external preprocessing.

\textbf{InternVL3: Native Multimodal Pre-training at Scale.}
InternVL3 \cite{zhu2025internvl3} departs from the dominant “language-first, vision-second” paradigm by performing native multimodal pre-training from the outset: a single unified stage interleaves large-scale text corpora with diverse vision–language data, allowing linguistic and visual capacities to co-evolve without later alignment.
Architecturally, the model keeps the clean ViT–MLP–LLM stack: an image-agnostic ViT feeds a lightweight two-layer projector that directly maps raw visual tokens into the language model’s embedding space, eliminating resolution-specific modules or fixed-size crops.
To gracefully accommodate long visual contexts, InternVL3 adopts Variable Visual Position Encoding (V2PE) \cite{ge2025v2pe} that modulates positional increments for visual tokens, keeping the overall sequence length within the LLM’s native window.
Operationally, all components—vision encoder, projector, and language backbone—are jointly optimized on the combined corpus; subsequent post-training refines conversational quality through supervised fine-tuning followed by Mixed Preference Optimization (MPO) \cite{wang2024mpo}, a unified objective that blends preference, quality, and generation losses.
Test-time computation is amplified by a visual process-supervised critic that selects the best among multiple decoded chains, pushing reasoning quality without enlarging the model.
\begin{algorithm}[!tbp]
\caption{\method{} Pipeline}
\label{algorithm}
\KwIn{Video $V$; Question $I$; Target response $s$; Pretrained VLM; Pretrained VDM}
\KwOut{Fine-tuned VLM; Total loss $\mathcal{L}_{total}$}

\textbf{Step 1: VDM-based Attention Extraction}\\
$z_{\text{vision}} \gets \text{VDM.VAE.encode}(V)$ \tcp*{\text{Encode video}, $z_{\text{vision}}\in\mathbb{R}^{F\times H\times W}$}
$z_{\text{noise}} \gets \text{DDIM\_Inversion}(z_{\text{vision}}, \text{step}=5)$ \tcp*{Obtain DDIM-inverted noise}
$z_{\text{text}} \gets \text{VDM.TextEncoder}(I)$ \tcp*{Encode text}
$z_{\text{mm}} \gets [ z_{\text{text}}, z_{\text{noise}}]$ \tcp*{Form multimodal sequence}
\ForEach{DDIM denoising step}{
    Compute attention: \small$A_{\text{mm}} \gets \operatorname{Softmax}\left(\frac{Q_{\text{mm}}K_{\text{mm}}^T}{\sqrt{d}}\right)$;\\
    Average $A_{\text{mm}}$ for all layers;\\
    Store $A_{\text{mm}}$ for all heads;
}

\textbf{Step 2: Motion-sensitive Head Selection}\\
\ForEach{head $h$ in VDM}{
    Extract vision-to-vision attention $A^{(h)}_{\text{v2v}}$ from $A_{\text{mm}}$ \tcp*{$[S\times S], S=H\times W$}
    Compute DFC, TCS, and DSR
}
$\text{score}_h = \text{Norm(DFC)} + \text{Norm(TCS)} + \text{Norm(DSR)}$ \tcp*{Compute head score}
$A_{\text{mm}}^{\mathcal{H}_m} \gets \text{Avg}(\{A_{\text{mm}}^{(h)}\}_{h \in \mathcal{H}_m})$ \tcp*{Select top-$50\%$ heads $\mathcal{H}_m$, and aggregate motion heads}

\textbf{Step 3: Motion-salient Text Token Identification}\\
$A_{\text{t2v}} \gets \text{extract text-to-vision region}(A_{\text{mm}}^{\mathcal{H}_m})$ \tcp*{Extract text-to-vision attention}
\ForEach{text token $t$}{
    $A_{t,\text{t2f}} \gets \text{Avg}_{(h,w)} A_{\text{t2v}}[t, f, h, w]$ \tcp*{Pool over spatial dimensions}
    $m_t \gets \text{Mean}_f(A_{t,\text{t2f}})$ \tcp*{Compute mean}
    $d_t \gets \frac{1}{F-1} \sum_{f=1}^{F-1} |A_{t,\text{t2f}}[f+1] - A_{t,\text{t2f}}[f]|$ \tcp*{Compute first-order diff mean}
    $MS(t) = m_t + d_t$ \tcp*{Motion score}
}
Select top-$50$\% tokens $\mathcal{T}_m$ by $MS(t)$\\
$A_{\text{VDM}} \gets A_{\text{t2v}}[\mathcal{T}_m, :]$ \tcp*{Obtain VDM motion prior}

\textbf{Step 4: VLM Attention Extraction}\\
Extract VLM attention maps $A_{\text{VLM,raw}}$ for $(V,I,s)$\\
Average pool across heads/layers and obtain $A_{\text{VLM}}$\\
$A_{\text{VLM}} \gets A_{\text{VLM}}[\mathcal{T}_m, :]$ \tcp*{Select same tokens}
Interpolate $A_{\text{VLM}}$ to match $A_{\text{VDM}}$

\textbf{Step 5: Supervised Fine-tuning with Attention Alignment}\\
$A_{\text{VLM,aligned}} \gets \text{MLP}(A_{\text{VLM}})$ // {Align VLM attention}
$\mathcal{L}_{\text{MSE}} = \|A_{\text{VLM,aligned}} - A_{\text{VDM}}\|_2$ \tcp*{Compute attention alignment loss}
$\mathcal{L}_{\text{AR}} = -\sum_{i=1}^{|s|} \log P(s[i] | V, I, \theta, s[1:i-1])$ \tcp*{Compute auto-regressive loss}
$\mathcal{L}_{\text{total}} = \mathcal{L}_{\text{AR}} + \lambda \mathcal{L}_{\text{MSE}}$ \tcp*{Compute total loss}
Update trainable parameters in VLM with $\mathcal{L}_{\text{total}}$
\end{algorithm}

\section{Algorithm}

The \method{} pipeline is shown in Algorithm \ref{algorithm}.
\textbf{(1) Step 1: VDM-based Attention Extraction.}
The input video is encoded into latent representations using a pre-trained VDM.
During the DDIM inversion and denoising process, multi-head attention maps are extracted from the VDM across all layers.
\textbf{(2) Step 2: Motion-sensitive Head Selection.}
For each attention head, motion relevance is quantitatively evaluated using the Diagonal Focus Coefficient, Temporal Continuity Score, and Diagonal Saliency Ratio.
The top-50\% heads most sensitive to temporal motion are selected, and their attention maps are aggregated.
\textbf{(3) Step 3: Motion-salient Text Token Identification.}
The aggregated attention maps are analyzed to compute temporal attention statistics for each text token.
Tokens exhibiting high temporal attention and significant inter-frame variation are selected as motion-salient text tokens.
The VDM attention maps filtered by both MHS and MTTI are used to construct motion priors, focusing on motion-relevant regions and tokens.
\textbf{(4) Step 4: VLM Attention Extraction.}
For the same video-text input, attention maps are extracted from the VLM and pooled across heads and layers.
The attention corresponding to the selected motion-salient tokens is retained and interpolated if necessary.
\textbf{(5) Step 5: Supervised Fine-tuning with Attention Alignment}
During fine-tuning, the VLM is optimized by minimizing the L2 distance between its own attention maps and the motion priors from the VDM, while also being supervised with the standard auto-regressive loss for video question answering.
All trainable parameters in the VLM are updated with respect to the combined loss, enabling the model to acquire enhanced motion understanding without modifying its architecture or adding new trainable modules.

\section{Detailed Theoretical Analysis}

This section provides a comprehensive theoretical extension from the main text, offering deeper insights into the distributional properties of VLMs and VDMs that form the foundation of \method{}.

\subsection{Preliminaries and Notation}

Let a video be denoted as
\begin{equation}
    V = (V_1, V_2, \dots, V_F),
\end{equation}
where $V_f$ is the $f$-th frame and $F$ is the number of frames.  
Each frame is discretized into spatial locations $s \in \mathcal{S}$; we denote the visual feature at location $s$ and frame $f$ as $v_{s,f}$.

A VLM is trained with an autoregressive objective:
\begin{equation}
    \mathcal{L}_{\mathrm{AR}}
    = -\sum_{i} \log p_{\theta}(r_i \mid V, r_{<i}),
\end{equation}
where $\theta$ denotes the VLM parameters and $r_i$ is the $i$-th response token.

Internally, the VLM produces text-to-video attention scores.  
For a token $t$, the attention weight at location $(s,f)$ can be interpreted as a discriminative conditional distribution:
\begin{equation}
    A^{\mathrm{VLM}}(t, s, f)
    \approx p_{\theta}\bigl(t \mid v_{s,f}, r_{<i}\bigr),
\end{equation}
which reflects a recognition-style distribution $p_{\theta}(t|V)$ learned under the autoregressive training paradigm.

By contrast, motion understanding requires discovering \emph{evidence} in the video that supports a motion-related token $t$ and its associated answer $a$. As argued in Sec.~3, this can be expressed via latent evidence $E$ and its factorization over semantic units $t \in T(q,a)$ as:
\begin{equation}
    p^\star(a \mid V, q) = \sum_{E} p^\star(a \mid E, q)\, p^\star(E \mid V, q),
\end{equation}
\begin{equation}
    E = \{E_t \mid t \in T(q,a)\},
\end{equation}
\begin{equation}
    p^\star(E \mid V, q) = \prod_{t \in T(q,a)} p^\star(E_t \mid V, t, q),
\end{equation}
where $E_t$ identifies the spatiotemporal regions that visually realize token $t$.

Within each $E_t$, the relevance of a location $(s,f)$ can be represented by a \emph{concept-conditioned visual likelihood}:
\begin{equation}
    r_t(s,f) = p^\star(v_{s,f} \mid t),
\end{equation}
which naturally leads to an \emph{evidence-seeking} distribution $p^\star(V \mid t)$: “given concept $t$, where and how is it visually realized over space and time?”

For motion-related tokens (e.g., verbs, motion modifiers), the latent evidence must encode \emph{frame-to-frame transitions}, which, at the feature level, can be summarized by a motion magnitude:
\begin{equation}
    \mathrm{Motion}(s,f) \propto \| V_{f+1}(s) - V_f(s) \|.
\end{equation}

\subsection{Distributional Mismatch in More Detail}

We now formalize the mismatch between $p_\theta(t \mid V)$ (what the VLM is trained on) and $p^\star(V \mid t)$ (what motion understanding requires).  
Using Bayes’ rule at the level of per-location features:
\begin{equation}
    p^\star(t \mid v_{s,f})
    = \frac{p^\star(v_{s,f} \mid t)\, p^\star(t)}{p^\star(v_{s,f})}.
\end{equation}

If the VLM were trained with sufficiently rich supervision and under a well-specified model, 
$A^{\mathrm{VLM}}(t,s,f)$ could in principle approximate $p^\star(t \mid v_{s,f})$.
However, in realistic settings, the supervision usually consists of high-level QA or captions rather than dense, per-location labels.
And the set of questions and responses $(q,a)$ is highly imbalanced: many queries can be answered from a static context without modeling temporal changes.
Also, the model capacity and optimization are shaped to maximize accuracy, not to invert the distribution to recover $p^\star(v_{s,f} \mid t)$.

Hence, a VLM can achieve high training and evaluation performance by relying mostly on static appearance cues.  
Formally, decompose the visual feature as:
\begin{equation}
    v_{s,f} = v^{\mathrm{stat}}_{s,f} + v^{\mathrm{dyn}}_{s,f},
\end{equation}
where $v^{\mathrm{stat}}$ captures static background and identity cues, and $v^{\mathrm{dyn}}$ captures temporal variations.
If most training questions can be answered using $v^{\mathrm{stat}}$ alone, then gradients of $\mathcal{L}_{\mathrm{AR}}$ with respect to $v^{\mathrm{dyn}}$ are small or sparse, and the model converges to a solution where:
\begin{equation}
    p_\theta(t \mid v_{s,f}) \approx p_\theta(t \mid v^{\mathrm{stat}}_{s,f}),
\end{equation}
i.e., the conditional depends only weakly on temporal dynamics.

On the other hand, motion understanding requires estimating where and when motion-related evidence occurs, which is precisely governed by $p^\star(v_{s,f} \mid t)$ and its temporal variation across $f$. 
In particular, the expectation
\begin{equation}
    \mathbb{E}\bigl[\mathrm{Motion}(s,f) \mid t\bigr]
    \propto 
    \mathbb{E}\Bigl[\bigl\| V_{f+1}(s) - V_f(s) \bigr\| \,\Big|\, t \Bigr]
\end{equation}
is a functional of the evidence-seeking distribution $p^\star(V \mid t)$, not of the recognition distribution $p^\star(t \mid V)$.

Thus, unless the training signal explicitly forces the VLM to approximate or invert $p^\star(V \mid t)$, there is a \textbf{structural mismatch}: the model is optimized for predicting tokens given videos, while fine-grained motion reasoning requires discovering the video given a concept. 
This explains why VLMs exhibit appearance bias and weak temporal sensitivity, as observed empirically.

\subsection{VDM Attention as Evidence-Seeking Posterior}

Video diffusion models (VDMs) are trained to generate videos conditioned on text. 
At the denoising step $k$, the model receives noisy latent video $z_k$ and text tokens $t$, and outputs a cleaner latent that approximates the true video latent $z_0$. 
The cross-attention at this step is computed as:
\begin{equation}
    A^{\mathrm{VDM}}(t, s, f)
    = \mathrm{Softmax}\!\left(
        \frac{Q(t)\, K(s,f)^{\top}}{\sqrt{d}}
    \right),
\end{equation}
where $Q(t)$ is the query for token $t$ and $K(s,f)$ is the key derived from the latent 
at location $(s,f)$. Aggregating across layers and timesteps, as described in the main paper, 
yields a stable attention estimate.

Under mild assumptions (e.g., linear attention maps and sufficiently expressive 
$Q/K$ projections), this attention can be interpreted as a \emph{normalized relevance score} 
between $t$ and $v_{s,f}$ conditioned on the current noisy latent:
\begin{equation}
    A^{\mathrm{VDM}}(t, s, f)
    \approx 
    p_{\phi}(v_{s,f} \mid t, z_k),
\end{equation}
where $\phi$ denotes VDM parameters. This resembles the 
concept-conditioned likelihood $p^\star(v_{s,f}|t)$ required for motion reasoning.

Two properties make VDM attention particularly suited as a motion prior:
\textbf{Evidence-seeking nature.}
Since the model must generate video content that visually realizes the text token $t$, the attention highlights locations where the semantics of $t$ should appear. 
This approximates the evidence-seeking distribution $p^\star(V|t)$: regions receiving higher attention are those where the generated video is more likely to exhibit the concept.
\textbf{Motion calibration.}
The denoising objective penalizes reconstruction errors, which are larger in regions with strong motion (large $\| V_{f+1}(s) - V_f(s) \|$) because these regions are harder to predict from noisy latents. 
Consequently, gradients concentrate on dynamic areas, making cross-attention for motion-related tokens sensitive to temporal variations. 
Locations with higher motion magnitude receive higher or more fluctuating attention across timesteps, aligning with the notion of motion evidence.

In the idealized limit where the VDM perfectly models the true data distribution,
$A^{\mathrm{VDM}}(t, s, f)$ converges to a calibrated approximation of 
$p^\star(v_{s,f} \mid t)$. Even in practice, it serves as a high-quality, 
motion-aware prior that complements the VLM's appearance-focused attention.

\subsection{Attention Alignment as Approximate Posterior Matching}

\method{} introduces an auxiliary MSE loss between VDM and VLM attention:
\begin{equation}
    \mathcal{L}_{\mathrm{MSE}}
    = \bigl\| \mathrm{Aligner}(A^{\mathrm{VLM}}) - A^{\mathrm{VDM}} \bigr\|_2^{2},
\end{equation}
and optimizes
\begin{equation}
    \mathcal{L}_{\mathrm{total}}
    = \mathcal{L}_{\mathrm{AR}} + \lambda \,\mathcal{L}_{\mathrm{MSE}}.
\end{equation}

For a given motion-salient token $t$, we treat each attention row as a discrete 
distribution over locations:
\begin{equation}
    \pi_{\theta}^{\mathrm{VLM}}(s,f \mid t)
    := A^{\mathrm{VLM}}(t,s,f),
\end{equation}
\begin{equation}
    \pi_{\phi}^{\mathrm{VDM}}(s,f \mid t)
    := A^{\mathrm{VDM}}(t,s,f).
\end{equation}

Ignoring the Aligner for a moment (or assuming it is near identity), we may view $\mathcal{L}_{\mathrm{MSE}}$ as minimizing an $L^{2}$ distance between these distributions.
In expectation over data, we approximately solve:
\begin{equation}\label{eq24}
\begin{aligned}
    \min_{\theta}\quad &
    \mathbb{E}_{(V,q,a)}
    \bigl[ -\log p_{\theta}(a \mid V,q) \bigr]+
    \\
    \lambda \,
    &\mathbb{E}_{t \in T_{\mathrm{mot}}(q,a)}
    \Bigl[
        \bigl\|
            \pi_{\theta}^{\mathrm{VLM}}(\cdot \mid t)
            -
            \pi_{\phi}^{\mathrm{VDM}}(\cdot \mid t)
        \bigr\|_{2}^{2}
    \Bigr].
\end{aligned}
\end{equation}

This expanded objective has two important interpretations:
\textbf{Posterior matching in attention space.}
If we interpret 
$\pi_{\theta}^{\mathrm{VLM}}(\cdot \mid t)$ 
as an approximate posterior over evidence locations for token $t$, 
and 
$\pi_{\phi}^{\mathrm{VDM}}(\cdot \mid t)$ 
as an approximate concept-conditioned distribution from the generative model, 
then $\mathcal{L}_{\mathrm{MSE}}$ acts to push the VLM distribution toward the 
motion-sensitive evidence-seeking distribution encoded by the VDM. 
This reduces the discrepancy between the recognition distribution 
(where a VLM can rely heavily on static cues) 
and the generative distribution that better respects motion.
\textbf{Constrained optimization and regularization.}
The optimization in Eq. \ref{eq24} can equivalently be viewed as:
\begin{equation}
\begin{aligned}
    \min_{\theta}\quad 
        &\mathbb{E}\bigl[-\log p_{\theta}(a \mid V,q)\bigr]
    \\
    \text{s.t.}\quad
        &\mathbb{E}
            \bigl[
                \|\pi_{\theta}^{\mathrm{VLM}}
                - \pi_{\phi}^{\mathrm{VDM}}\|_{2}^{2}
            \bigr]
        \le \epsilon(\lambda),
\end{aligned}
\end{equation}
where $\epsilon(\lambda)$ decreases as $\lambda$ increases.
Thus, the feasible set of VLM parameters is restricted to those producing attention maps that are compatible with motion priors distilled from the VDM.

In practice, the Aligner is implemented as a small MLP. 
Its limited capacity enables mild, smooth transformations (e.g., rescaling or slight warping) to account for architectural differences between VLM and VDM, while preventing degenerate solutions in which the VLM attention remains unchanged and the Aligner absorbs the full discrepancy. 
In this sense, the Aligner functions as a preconditioner rather than a substitute for matching attention distributions.

\subsection{MHS and MTTI as Projection onto a Motion Subspace}

Both MHS and MTTI modules act as selection operators that isolate the motion-relevant components of the VDM attention. 
Let $\mathcal{A}$ denote the full text-to-video attention tensor from the VDM.

\textbf{Head-level projection.}  
MHS selects a subset of heads exhibiting strong temporal structure.
At a high level, this can be written as a projection:
\begin{equation}
    A_{\mathrm{head}} = \Pi_{\mathrm{head}}(\mathcal{A}),
\end{equation}
where $\Pi_{\mathrm{head}}$ keeps only motion-sensitive heads and discards others.

\textbf{Token-level projection.}  
Similarly, MTTI selects text tokens whose attention varies across frames. 
This yields a second projection:
\begin{equation}
    A_{\mathrm{mot}} = \Pi_{\mathrm{token}}(A_{\mathrm{head}}).
\end{equation}

\textbf{Overall effect.}
Together, MHS and MTTI produce a compact motion-focused attention map:
\begin{equation}
    A_{\mathrm{mot}} 
    = \Pi_{\mathrm{token}}\!\bigl(\Pi_{\mathrm{head}}(\mathcal{A})\bigr),
\end{equation}
which represents the projection of the original VDM attention onto a motion subspace. This projected attention is the signal aligned with the VLM in \method{}.

\begin{table}[!t]
\setlength{\tabcolsep}{0.8mm}
\caption{The experimental system and hardware setups.}
\centering
\begin{tabular}{cc} 
\toprule
\multicolumn{2}{c}{\textbf{System \& Hardware Overview}}  \\ 
\midrule
\multirow{2}{*}{CPU} & Intel(R) Xeon(R) Platinum~             \\
                     & 8375C CPU @ 2.90GHz                    \\
GPU                  & 8$\times$NVIDIA A100 Tensor Core GPU                             \\
Memory               & 1T  DRAM                                 \\
Operating System     & Ubuntu 22.04.4 LTS                     \\
CUDA Version         & 12.1                                   \\
NVIDIA Driver        & ~530.30.02                             \\
ML Framework         & Python 3.10.12  Pytorch 2.5.1          \\

\midrule
\multicolumn{2}{c}{\textbf{GPU Specifications}}  \\ 
\midrule
CUDA Cores& 6912 \\
Memory Capacity& 80GB \\
Memory Bandwidth& 1935GB/s \\
\bottomrule
\end{tabular}
\label{tab: hardware}
\end{table}

\begin{table}[!t]
\setlength{\tabcolsep}{1.0mm}
\caption{Statistical information of the two used benchmarks, including each task type and quantity.}
\centering
\begin{tabular}{c|cccccc} 
\toprule
\multirow{2}{*}{MotionBench} & MR   & LM   & AO   & RC   & MO   & CM   \\
                             & 1478 & 546  & 519  & 400  & 690  & 385  \\ 
\hline
\multirow{2}{*}{FAVOR-Bench} & AS   & HAC  & SAD  & MAD  & CM   & NSM  \\
                             & 2637 & 1541 & 1662 & 1205 & 1075 & 64   \\
\bottomrule
\end{tabular}
\label{Tab:statistic}
\end{table}

\section{Discussion}

\subsection{Dependence on VDM quality}

\method{}'s effectiveness depends on the informativeness of the VDM’s motion priors.
As discussed in the main content, when a VDM produces diffuse attention (e.g., large static subjects), the alignment signal weakens.
Nevertheless, using such priors can yield improvements, and our paper’s focus is on how to transfer those priors rather than on improving their intrinsic quality.
To mitigate this, we will employ solutions like confidence gating (low-motion-confidence cases get minimal alignment) or ensemble attention maps from multiple VDMs to reduce model-specific artifacts.

\subsection{Role of $\lambda$ in Alignment}

Tuning $\lambda$ effectively trades off trust in the VDM prior versus the VLM’s own training signal.
$\lambda=0$ means no prior injection and standard VLM.
Moderate $\lambda$ presents motion-sensitive heads and tokens are guided by priors, while static reasoning remains mostly unaffected.
Very large $\lambda$ denotes that VLM may overfit the VDM’s specific biases, potentially harming robustness.

\begin{table*}
\setlength{\tabcolsep}{0.9mm}
\centering
\caption{Quantitative results of MotionBench. * denotes results we reproduced using their open-source code, while other results are taken from the original benchmark.}
\begin{tabular}{lc|cc|cccccc} 
\toprule
Model                                                   & Frames & Overall                     & Average                     & MR    & LM    & CM    & MO    & AO    & RC     \\ 
\hline
\multicolumn{10}{l}{Small Size Series}                                                                                                                                                \\ 
\hline
Qwen2.5-VL-3B* \cite{bai2025qwen2}                                         & 1fps   & 53.56                       & 49.45                       & 59.54 & 53.11 & 38.44 & 70.14 & 40.46 & 35.00  \\
\textbf{Qwen2.5-VL-3B + MotionEnhancer (Ours)}          & 1fps   & $\text{56.60}^{\uparrow3.04}$ & $\text{52.51}^{\uparrow3.06}$ & 63.06 & 61.72 & 47.01 & 68.84 & 43.16 & 31.25  \\
InternVL3-2B* \cite{zhu2025internvl3}                                          & 8      &     53.96                        &    49.69                         &   60.01    &  57.69     &  43.90     &   70.00    &   40.27    &   26.25     \\
\textbf{InternVL3-2B + MotionEnhancer (Ours)} & 8      &            $\text{55.50}^{\uparrow1.54}$                 &          $\text{51.35}^{\uparrow1.66}$                  &  61.57     &   57.51    &   46.23    &   71.30    &  42.00     &   29.50     \\ 
\hline
\multicolumn{10}{l}{Medium Size Series}                                                                                                                                                \\ 
\hline
MiniCPM-V2.6-7B \cite{yao2024minicpm}                                        & 64     & 52                          & -                           & -     & -     & -     & -     & -     & -      \\
CogVLM2-Video-8B \cite{hong2024cogvlm2}                                       & 24     & 41                          & -                           & -     & -     & -     & -     & -     & -      \\
GLM4-9B + TE Fusion \cite{hong2025motionbench}                                    & 16     & 58                          & -                           & -     & -     & -     & -     & -     & -      \\
Qwen2.5-VL-7B* \cite{bai2025qwen2}                                         & 1fps   & 52.81                       & 48.29                       & 59.00 & 54.58 & 35.58 & 71.30 & 38.54 & 30.75  \\
Qwen2.5-VL-7B + MotionSight* \cite{du2025motionsight}                           & 1fps   & 55.30                       & 51.56                       & 59.88 & 57.33 & 47.01 & 73.91 & 40.46 & 30.75  \\
\textbf{Qwen2.5-VL-7B + MotionEnhancer (Ours)}          & 1fps   & $\text{57.04}^{\uparrow4.23}$ & $\text{52.92}^{\uparrow4.63}$ & 63.40 & 61.54 & 47.27 & 70.29 & 43.55 & 31.50  \\
InternVL3-8B* \cite{zhu2025internvl3}                                          & 8      & 54.88                       & 50.81                       & 60.42 & 58.06 & 43.64 & 70.29 & 43.93 & 28.50  \\
\textbf{InternVL3-8B +MotionEnhancer (Ours)} \cite{zhu2025internvl3}                                          & 8      & $\text{57.69}^{\uparrow2.81}$                       & $\text{53.22}^{\uparrow2.41}$                       & 64.14 & 60.07 & 48.83 & 75.94 & 40.85 & 29.50  \\
\hline
\multicolumn{10}{l}{Large Size Series}                                                                                                                                            \\ 
\hline
PLLaVA-34B \cite{xu2024pllava}                                             & 16     & 52                          & -                           & -     & -     & -     & -     & -     & -      \\
LLaVA-NeXT-Video-34B \cite{li2024llavanext}                                   & 32     & 48                          & -                           & -     & -     & -     & -     & -     & -      \\
Qwen2.5-VL-72B* \cite{bai2025qwen2}                                        & 1fps   & 58.30                       & 54.32                       & 64.00 & 60.30 & 48.60 & 73.20 & 46.80 & 33.00  \\
\bottomrule
\end{tabular}
\label{Tab:MotionBench}
\end{table*}

\begin{table*}[!t]
\setlength{\tabcolsep}{0.9mm}
\centering
\caption{Quantitative results of FAVOR-Bench. * denotes results we reproduced using their open-source code, while other results are taken from the original benchmark.}
\begin{tabular}{lc|cc|cccccc} 
\toprule
Model                         & Frames & Overall & Average & AS    & HAC   & SAD   & MAD    & CM     & NSM    \\ 
\hline
\multicolumn{10}{l}{Small Size Series}                                                                                \\ 
\hline
VideoLLaMA3-2B \cite{zhang2025videollama3}               & 1fps   & 32.98~  & 34.61   & 28.97 & 36.60 & 34.90 & 38.01  & 28.56  & 40.62  \\
InternVL2.5-2B \cite{chen2024internvl2.5}               & 8      & 22.90   & 23.45   & 18.70 & 28.23 & 23.71 & ~27.47 & ~19.16 & 23.44  \\
Qwen2.5VL-3B* \cite{bai2025qwen2}                 & 1fps   & 37.43~  & 38.07   & 38.45 & 38.16 & 39.35 & 43.40  & 23.72  & 45.31  \\
\textbf{Qwen2.5VL-3B + MotionEnhancer (Ours)} & 1fps   & $\text{44.53}^{\uparrow7.10}$   & $\text{43.94}^{\uparrow5.87}$   & 45.01 & 51.59 & 44.40 & 48.96  & 28.37  & 45.31  \\ 
InternVL3-2B* \cite{zhu2025internvl3}                 & 1fps   & 39.27~  & 39.11   & 37.66 & 43.28 & 40.49 & 44.98  & 29.21  & 39.06  \\
\textbf{InternVL3-2B + MotionEnhancer (Ours)} & 1fps   & $\text{43.71}^{\uparrow4.44}$   & $\text{45.35}^{\uparrow6.24}$   & 38.53 & 54.57 & 42.60 & 51.78  & 33.02  & 51.56  \\ 
\hline
\multicolumn{10}{l}{Medium Size Series}                                                                                \\ 
\hline
LLaVA-Video-7B-Qwen2 \cite{zhang2024llavavideo}         & 64     & 38.60   & 39.94   & 36.14 & 41.27 & 41.28 & 44.48  & 29.58  & 46.88  \\
VideoChat-Flash-Qwen2-7B \cite{li2024videochat}     & 1fps   & 43.82   & 44.86   & 41.90 & 48.41 & 42.84 & 50.95  & 35.07  & 50.00  \\
VideoLLaMA3-7B \cite{zhang2025videollama3}               & 1fps   & 41.46   & 41.46   & 40.20 & 44.13 & 42.42 & 48.30  & 31.53  & 42.19  \\
Video-LLaVA-7B \cite{lin2023videollava}               & 8      & 25.37   & 25.09   & 24.91 & 21.54 & 25.45 & 30.54  & 26.23  & 21.88  \\
LLaVA-NeXT-Video-7B \cite{li2024llavanext}          & 8      & 23.45   & 22.27   & 21.27 & 22.45 & 26.05 & 26.72  & 23.07  & 14.06  \\
Tarsier-7B \cite{wang2024tarsier}                   & 8      & 17.46   & 20.50   & 12.55 & 21.16 & 17.87 & 17.93  & 22.23  & 31.25  \\
Qwen2.5VL-7B* \cite{bai2025qwen2}                 & 1fps   & 42.61   & 42.58   & 41.64 & 47.83 & 44.89 & 47.55  & 28.28  & 45.31  \\
Qwen2.5VL-7B + MotionSight* \cite{du2025motionsight}  & 1fps   & 45.47   & 45.99   & 46.23 & 51.59 & 45.01 & 50.04  & 29.95  & 53.12  \\
\textbf{Qwen2.5VL-7B + MotionEnhancer (Ours)} & 1fps   & $\text{46.88}^{\uparrow4.27}$   & $\text{47.01}^{\uparrow4.43}$   & 49.34 & 50.62 & 45.37 & 53.20  & 30.42  & 53.12  \\ 
InternVL3-8B* \cite{zhu2025internvl3}                 & 1fps   & 45.82   & 46.35   & 45.39 & 48.54 & 47.59 & 51.45  & 33.58  & 51.56  \\
\textbf{InternVL3-8B + MotionEnhancer (Ours)} & 1fps   & $\text{48.94}^{\uparrow3.12}$   & $\text{49.25}^{\uparrow2.90}$   & 47.17 & 57.11 & 46.57 & 56.35  & 36.74  & 51.56  \\ 
\hline
\multicolumn{10}{l}{Large Size Series}                                                                            \\ 
\hline
LLaVA-NeXT-Video-34B \cite{li2024llavanext}         & 8      & 30.44   & 32.58   & 31.70 & 31.99 & 32.31 & 22.99  & 29.58  & 46.88  \\
LLaVA-Video-72B-Qwen2 \cite{zhang2024llavavideo}        & 64     & 46.08   & 46.49   & 48.35 & 47.50 & 45.25 & 51.70  & 33.02  & 53.12  \\
Qwen2.5-VL-72B* \cite{bai2025qwen2}               & 1fps   & 48.14   & 48.17   & 50.28 & 46.98 & 48.13 & 51.78  & 40.28  & 51.56  \\
InternVL2.5-78B \cite{chen2024internvl2.5}              & 8      & 38.54   & 38.36   & 38.38 & 40.62 & 39.05 & 43.65  & 29.40  & 39.06  \\
\bottomrule
\end{tabular}
\label{Tab:FAVOR-Bench}
\end{table*}

\begin{figure*}
    \centering
    \includegraphics[width=1.0\linewidth]{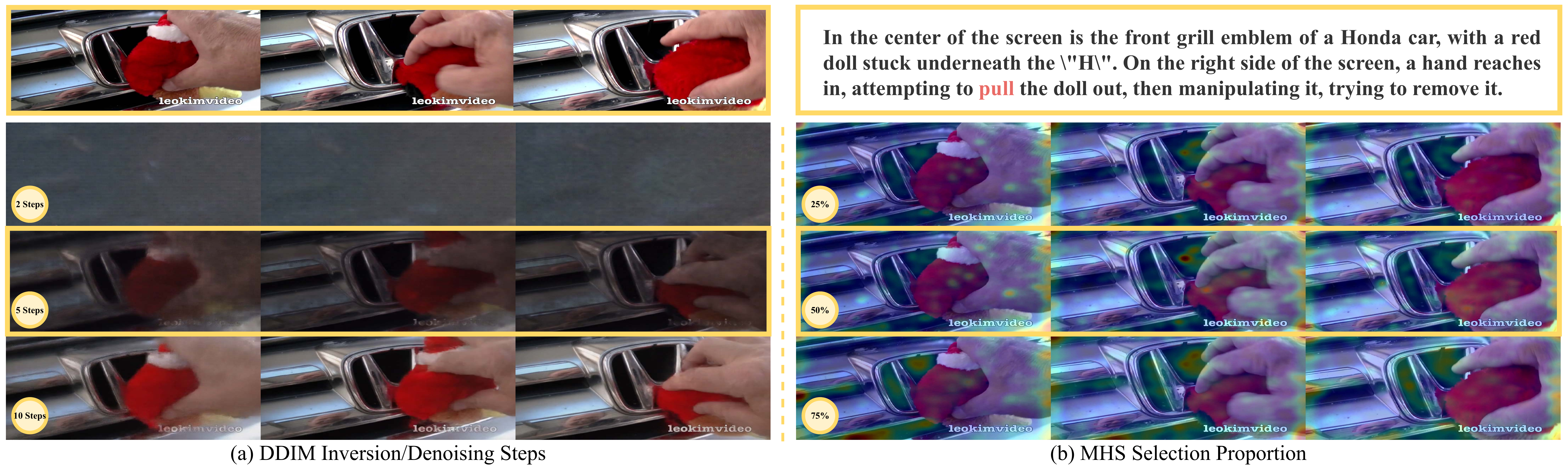}
    \caption{Visualization of different DDIM inversion/denoising steps and MHS selection proportion (left top: original video).}
    \label{fig:ablationcase}
\end{figure*}

\section{Experiments}

\subsection{System \& Hardware Setups}

The system and hardware details with GPU specifications of our experimental setup are provided in Tab.~\ref{tab: hardware}.

\subsection{Benchmark Details}

\textbf{Basic information.}
In our experiments, we select two public benchmarks, namely MotionBench \cite{hong2025motionbench} and FAVOR-Bench \cite{tu2025favorbench}, containing questions of different types from different video sources.
They have been used in previous VLM for video understanding studies, making them suitable for evaluating the performance of our method.
The details of these benchmarks are as follows:
\begin{itemize}
    \item \textbf{MotionBench} is a benchmark with 5,385 videos and a total of 8,052 question-answer pairs.
    It is divided into a development set with 4,018 samples and a test set with 4,036 samples.
    MotionBench evaluates six core capabilities for understanding fine details in motion: Motion Recognition (MR), Location-related Motion (LM), Action Order (AO), Repetition Count (RC), Motion-related Objects (MO), and Camera Motion (CM).
    This allows for a comprehensive evaluation of motion-level perception.
    The videos come from diverse sources, including the web, existing public datasets, and synthetic videos created using Unity3.
    This mix ensures broad coverage of real-world applications.
    All data underwent careful human annotation and a multi-stage quality control process.
    
    \item \textbf{FAVOR-Bench} is a benchmark of 1,776 carefully selected videos covering diverse domains, each with detailed manual annotations of various motions.
    It evaluates models through both close-ended and open-ended tasks.
    For close-ended evaluation, it contains 8,184 challenging question-answer pairs across six tasks: Action Sequence (AS), Holistic Action Classification (HAC), Single Action Detail (SAD), Multiple Action Details (MAD), Camera Motion (CM), and Non-Subject Motion (NSM).
    For open-ended evaluation, FAVOR-Bench offers both a novel cost-efficient LLM-free caption assessment method and a GPT-assisted evaluation approach.
\end{itemize}

\noindent \textbf{Benchmark Statistics.}
We provide statistical information of the MotionBench dev set and FAVOR-Bench close-ended evaluation in Tab.~\ref{Tab:statistic}, including each task type and its corresponding quantity.

\subsection{Comparison with More VLMs}

Given the page constraint of the main text, the comparative results of additional VLMs on the FAVOR-Bench benchmark are presented in Tab.~\ref{Tab:MotionBench} and Tab.~\ref{Tab:FAVOR-Bench}.
Empowered by \method{}, Qwen2.5-VL and InternVL3 not only outperform their counterparts within the same model size category but also attain performance comparable to that of larger-scale VLMs.

\subsection{More Ablations}

We conduct ablation studies on DDIM inversion/denoising steps and MHS/MTTI selection ratio.
Tab.~\ref{Tab:ablation} shows that the number of DDIM steps has a clear impact on performance.
Using only 2 steps leads to obvious degradation on both benchmarks, especially on FAVOR-Bench, indicating that overly coarse inversion/denoising fails to recover reliable motion-sensitive attention from the VDM.
Increasing the number of steps consistently improves the results, as finer inversion and denoising better preserve temporally meaningful motion cues.
However, more steps also introduce higher computational cost.
The visual differences under different step settings are shown in Fig.~\ref{fig:ablationcase}(A), and we use 5 steps because the reconstructed dynamics are good for human eyes at low cost.

The attention head and text token selection ratio also affects performance by controlling the purity and coverage of the motion prior.
Fig.~\ref{fig:ablationcase}(B) and Tab.~\ref{Tab:ablation} show the visual and quantitative changes under different ratios.
A small ratio (25\%) makes the prior overly sparse and may discard useful motion heads or motion-related text tokens, resulting in incomplete motion supervision.
In contrast, a large ratio (75\%) introduces more non-motion attention, which weakens the motion signal and hurts alignment quality.
The 50\% setting gives the best trade-off between retaining sufficient motion information and filtering out irrelevant noise, leading to the most balanced overall performance.
Therefore, we use 50\% as a stable default.

\begin{table}[!t]
\setlength{\tabcolsep}{1.0mm}
\caption{Ablation of DDIM inversion/denoising steps and MHS/ MTTI selection ratio. \textbf{All results are trained with 25k data.}}
\centering
\begin{tabular}{l|cc|cc} 
\toprule
\multirow{2}{*}{Model Variant}        & \multicolumn{2}{c|}{MotionBench} & \multicolumn{2}{c}{FAVOR-Bench}  \\
                              & Over. & Aver.                & Over. & Aver.                \\ 
\hline
Qwen2.5VL-7B                  & 54.83     & 51.51                    & 44.83     & 44.54                    \\
- 2 steps for DDIM  & 52.76     & 48.68                    & 24.01     & 25.16                    \\
- 5 steps for DDIM  & 57.04     & 52.92                    & 46.88     & 47.01                    \\
- 10 steps for DDIM  & 57.51     & 53.35                    & 49.02     & 47.99                    \\
\hline
- 25\% head/token select & 55.33     & 51.00                    & 47.12     & 47.28                    \\
- 50\% head/token select & 57.04     & 52.92                    & 46.88     & 47.01                    \\
- 75\% head/token select & 54.88     & 50.69                    & 45.69     & 44.10                    \\
\bottomrule
\end{tabular}
\label{Tab:ablation}
\end{table}

\subsection{\method{} VS. More Training Data}

Tab.~\ref{Tab:moredata} analyzes the effectiveness of MotionEnhancer compared to direct data scaling on MotionBench and FAVOR-Bench using Qwen2.5-VL-7B as the baseline VLM.
Notably, MotionVid-QA \cite{du2025motionsight} is a large dataset containing 133k samples.
While trained solely on MotionVid-QA, we observe clear improvements over the baseline VLM (row 2).
Crucially, despite this substantial difference in training data volume, \method{} achieves superior performance on MotionBench (both Overall and Average scores) while delivering comparable performance on FAVOR-Bench with only 25k training data.
These results highlight \method{}'s ability to achieve competitive or superior outcomes with significantly less training data, underscoring its data efficiency and effectiveness in enhancing motion understanding for VLMs.

\subsection{Results on Conventional Benchmarks}

\textbf{SEED-Bench} \cite{li2023seed} is a large-scale benchmark with 19K human-annotated multiple-choice questions across 12 spatial and temporal dimensions on images and videos, designed to objectively evaluate the generative comprehension capabilities of multimodal large language models.
\textbf{Video-MME} \cite{fu2025videomme} is the first comprehensive benchmark for evaluating multimodal large language models on video understanding, featuring 900 manually curated videos across diverse domains and durations, with 2,700 expert-annotated multiple-choice questions.
It uniquely incorporates subtitles and audio to assess models' capabilities in handling complex, long-form, and multimodal video content.

As shown in Tab.~\ref{tab:VideoQA}, across both benchmarks, our method demonstrates that enhancing motion modeling does not come at the cost of traditional video understanding performance.
Instead, \method{} yields consistent improvements or preserves strong baselines, confirming its compatibility with general-purpose video comprehension tasks.
For Qwen2.5-VL-7B, \method{} slightly reduces the SeedBench “All” score, but improves the video subset from 61.3 to 62.1 and maintains competitive image-level performance.
On VideoMME, the model equipped with \method{} achieves an overall score of 61.4, while achieving competitive results on short, medium, and long video categories—demonstrating that temporal enhancements remain stable across video durations.
For InternVL3-8B, the trend is similarly positive, where \method{} preserves strong SeedBench results.
Importantly, on VideoMME, \method{} improves the overall score from 62.1 to 62.3, with consistent benefits across short and medium segments.
This suggests that \method{} strengthens temporal reasoning without materially harming static-image understanding.

Overall, these results indicate that \method{} not only boosts performance on motion-centric tasks but also maintains or even subtly improves general video understanding ability.
This demonstrates that injecting motion-aware priors does not introduce instability or unwanted trade-offs, reinforcing \method{}'s suitability as a lightweight and broadly applicable enhancement for video-language models.

\begin{table}[!t]
\setlength{\tabcolsep}{1.0mm}
\caption{Experimental results of \method{} VS. More Training Data on MotionBench and FAVOR-Bench using Qwen2.5-VL-7B as our backbone.}
\centering
\begin{tabular}{l|cc|cc} 
\toprule
\multirow{2}{*}{Model}        & \multicolumn{2}{c|}{MotionBench} & \multicolumn{2}{c}{FAVOR-Bench}  \\
                              & Over. & Aver.                & Over. & Aver.                \\ 
\hline
Qwen2.5VL-7B                  & 52.81     & 48.29                    & 42.61     & 42.58                    \\
+ MotionVid-QA (133k)  & 55.70     & 51.25                    & 47.12     & 47.28                    \\
+ MotionEnhancer (25k) & 57.04     & 52.92                    & 46.88     & 47.01                    \\
\bottomrule
\end{tabular}
\label{Tab:moredata}
\end{table}
\begin{table*}
\centering
\caption{Performance on SEED-Bench and VideoMME (w/o sub). * denotes results we reproduced using their open-source code, while other results are taken from the original benchmark.}
\begin{tabular}{l|ccc|cccc} 
\toprule
\multirow{2}{*}{Model}       & \multicolumn{3}{c|}{SEED-Bench} & \multicolumn{4}{c}{VideoMME (w/o sub)}   \\ 
\cline{2-8}
                             & All  & Image & Video           & Overall & Short & Medium & Long  \\ 
\hline
GPT-4V \cite{achiam2023gpt4}                      & 67.3 & 69.1  & 60.5            & 59.9    & 70.5  & 55.8   & 53.5  \\
LLaVA-1.5-13B \cite{liu2023llava}               & 61.6 & 68.2  & 42.7            & -       & -     & -      & -     \\
Weitu-VL-1.0-13B             & 69.2 & 74.2  & 50.5            & -       & -     & -      & -     \\
SPHINXv2-1k-13B \cite{lin2023sphinx}             & 67.5 & 74.8  & 39.8            & -       & -     & -      & -     \\
VideoLLaMA3-7B \cite{zhang2025videollama3}              & -    & -     & -               & 66.2    & 80.1  & 63.7   & 54.9  \\
VideoChat-Flash-7B \cite{li2024videochat}          & -    & -     & -               & 65.3    & 78.0  & 67.8   & 55.6  \\
VITA1.5-7B \cite{fu2025vita}                  & -    & -     & -               & 56.1    & 67.0  & 54.2   & 47.1  \\ 
\hline
Qwen2.5-VL-7B* \cite{bai2025qwen2}               & 74.1 & 77.5  & 61.3            & 60.7    & 70.9  & 59.6   & 51.6  \\
\textbf{Qwen2.5-VL-7B+MotionEnhancer (Ours)} & 72.9 & 75.9  & 62.1            & 61.4    & 72.8  & 61.1   & 50.2  \\
InternVL3-8B* \cite{zhu2025internvl3}                & 72.7 & 76.2  & 60.4            & 62.1    & 71.8  & 61.9   & 52.6  \\
\textbf{InternVL3-8B+MotionEnhancer (Ours)}  & 72.4 & 75.8  & 60.6            & 62.3    & 73.0  & 62.2   & 51.7  \\
\bottomrule
\end{tabular}
\label{tab:VideoQA}
\end{table*}

\section{Visulization Examples}

\subsection{A Complete Training Sample}

\begin{figure*}
    \centering
    \includegraphics[width=0.9\linewidth]{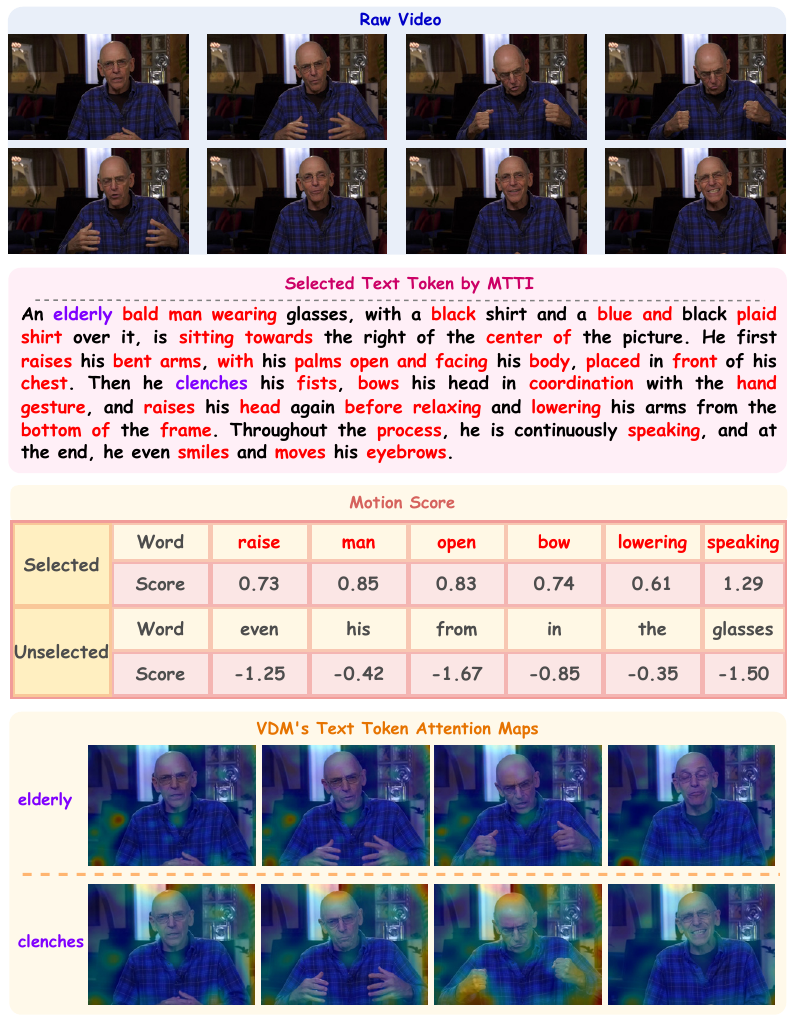}
    \caption{A complete training sample.
    We first show the selected text token by MTTI (in red).
    We then present the motion scores of some text tokens based on the mean value and the first-order difference mean.
    We further illustrate two aggregated text-to-vision attention maps.}
    \label{fig:attention}
\end{figure*}

To demonstrate the motion-aware token selection process, we provide a detailed visualization of a training instance in Fig.~\ref{fig:attention}.
The video depicts an elderly man performing a sequence of actions: raising his arms, clenching his fists, bowing his head, and speaking continuously.
These actions are annotated with descriptive text, from which we extract tokens using the MTTI.
Tokens such as “raise”, “bow”, “lowering”, and “speaking” receive high MS values, indicating strong correlation with motion dynamics.
In contrast, static or contextual tokens like “his”, “the”, or “glasses” are assigned low scores and are filtered out. This selection process ensures that only semantically and dynamically relevant tokens contribute to the motion prior.

\begin{figure}[t]
    \centering
    \includegraphics[width=1.0\linewidth]{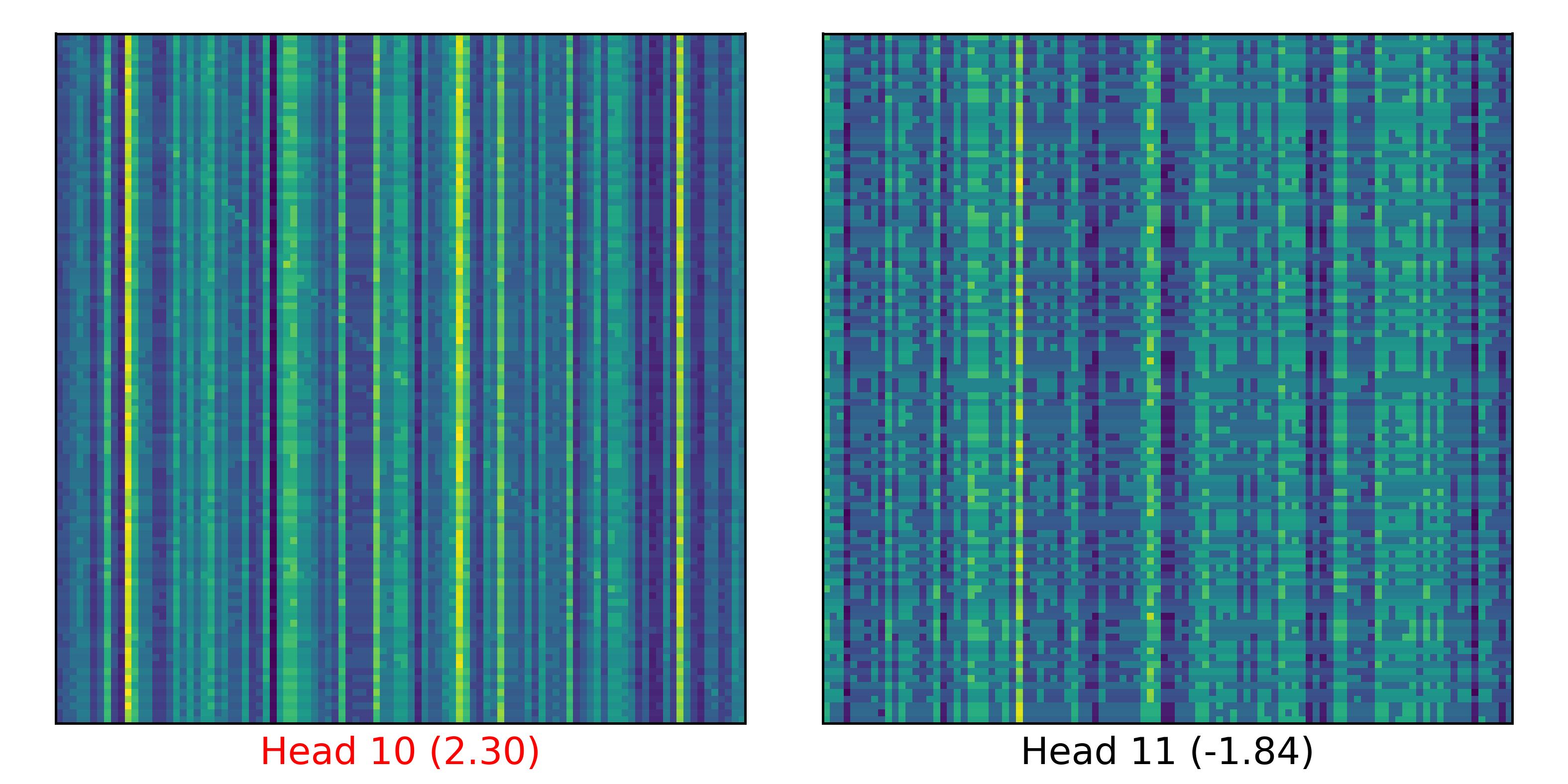}
    \caption{Zoom-in view of two specific heads.
    We selected the first 100 spatial location tokens for visualization.
    It is demonstrated that heads with high scores exhibit a diagonal pattern, which is consistent with the main paper.}
    \label{fig:specifichead}
\end{figure}

\begin{figure*}
    \centering
    \includegraphics[width=0.95\linewidth]{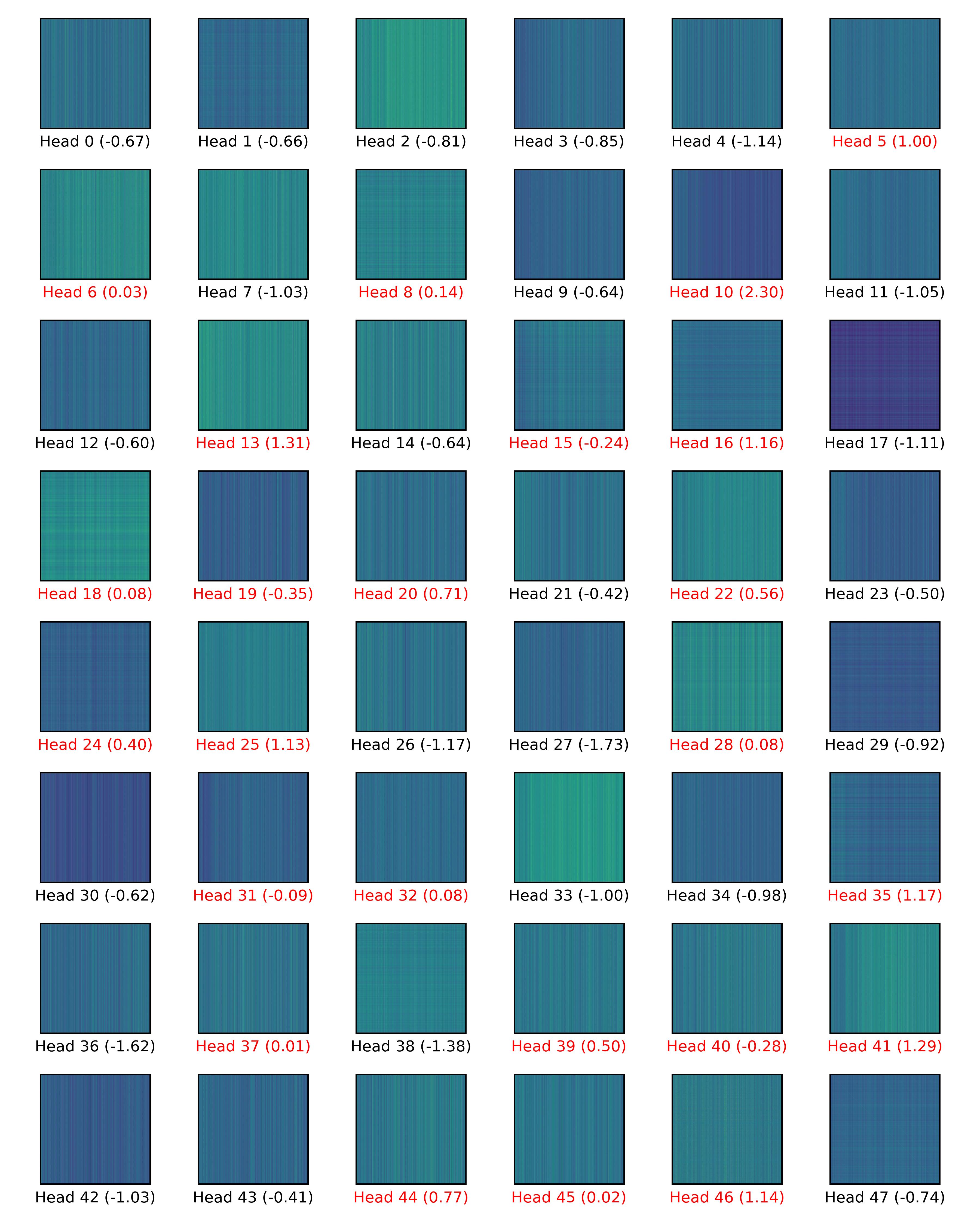}
    \caption{Vision-to-vision attention maps of different heads in VDM, with each head bearing a total score of DFC, TCS, and DSR.
    Heads in red are selected for aggregation.}
    \label{fig:attnheads}
\end{figure*}

Furthermore, we visualize the aggregated text-to-vision attention maps for the selected tokens in Fig.~\ref{fig:specifichead} and Fig.~\ref{fig:attnheads}.
These maps highlight spatial regions in the video frames that correspond to the described actions, confirming that the model attends to the correct locations where motion occurs.
This not only validates the effectiveness of MTTI but also provides interpretable evidence of how \method{} grounds language in dynamic visual contexts.

\subsection{DDIM Inversion and Reconstruction}

\begin{figure*}
    \centering
    \includegraphics[width=0.95\linewidth]{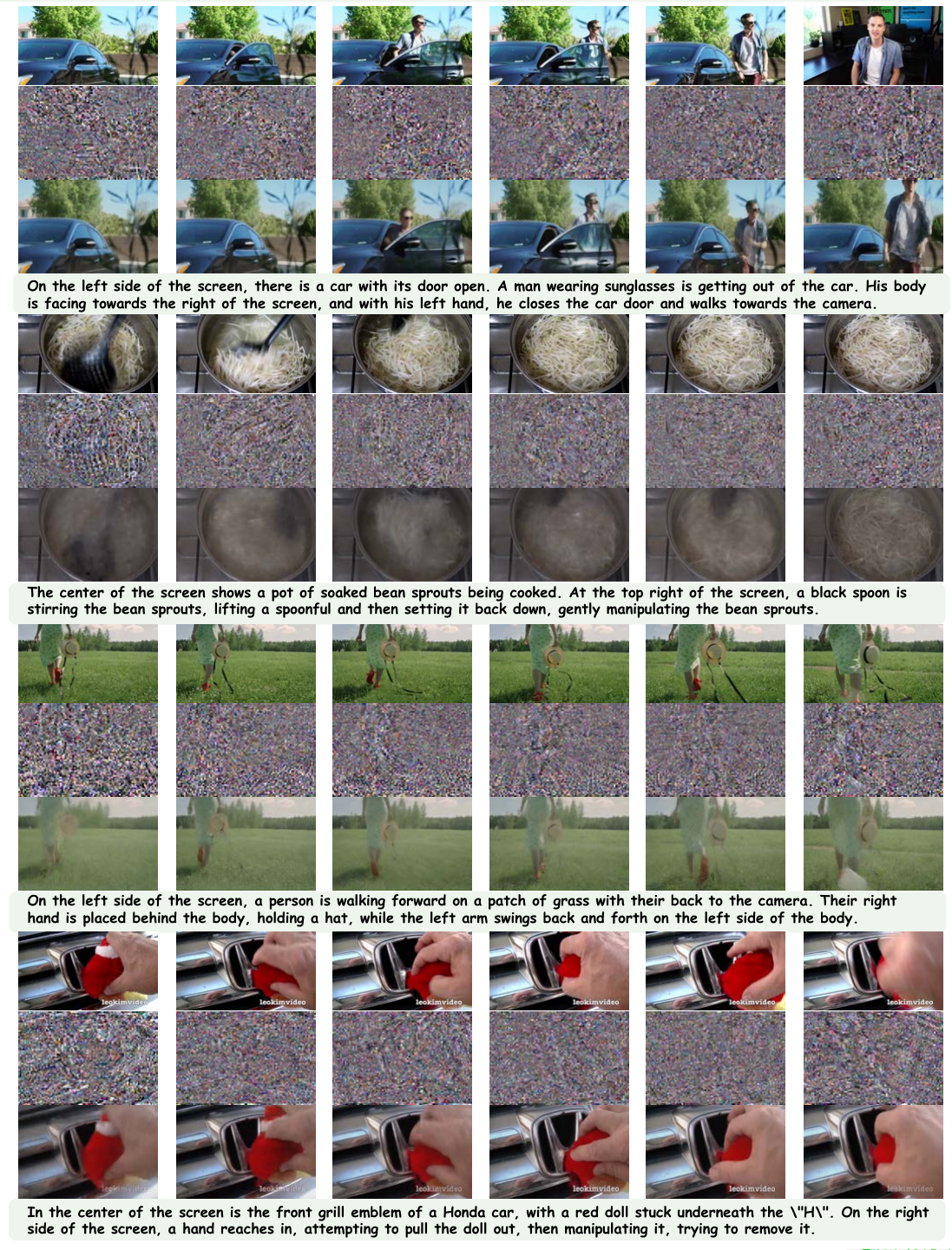}
    \caption{Our DDIM inversion and reconstruction process.
    Grouped in sets of three rows: the first row is the original video, the second row is the latent noise after DDIM Inversion, and the third row is the reconstructed video.
    }
    \label{fig:ddiminversion}
\end{figure*}

Fig.~\ref{fig:ddiminversion} illustrates the DDIM inversion and reconstruction pipeline, a core component of our attention extraction mechanism. The process begins with the original video frames, which are encoded into latent representations using the VDM’s VAE.
These latents are then inverted through a deterministic DDIM process to obtain a noise sequence that corresponds to the original video.
This reversibility is crucial for ensuring that the attention maps extracted during the denoising process are semantically aligned with the original video content.

\subsection{Video Understanding}

\begin{figure*}
    \centering
    \includegraphics[width=0.95\linewidth]{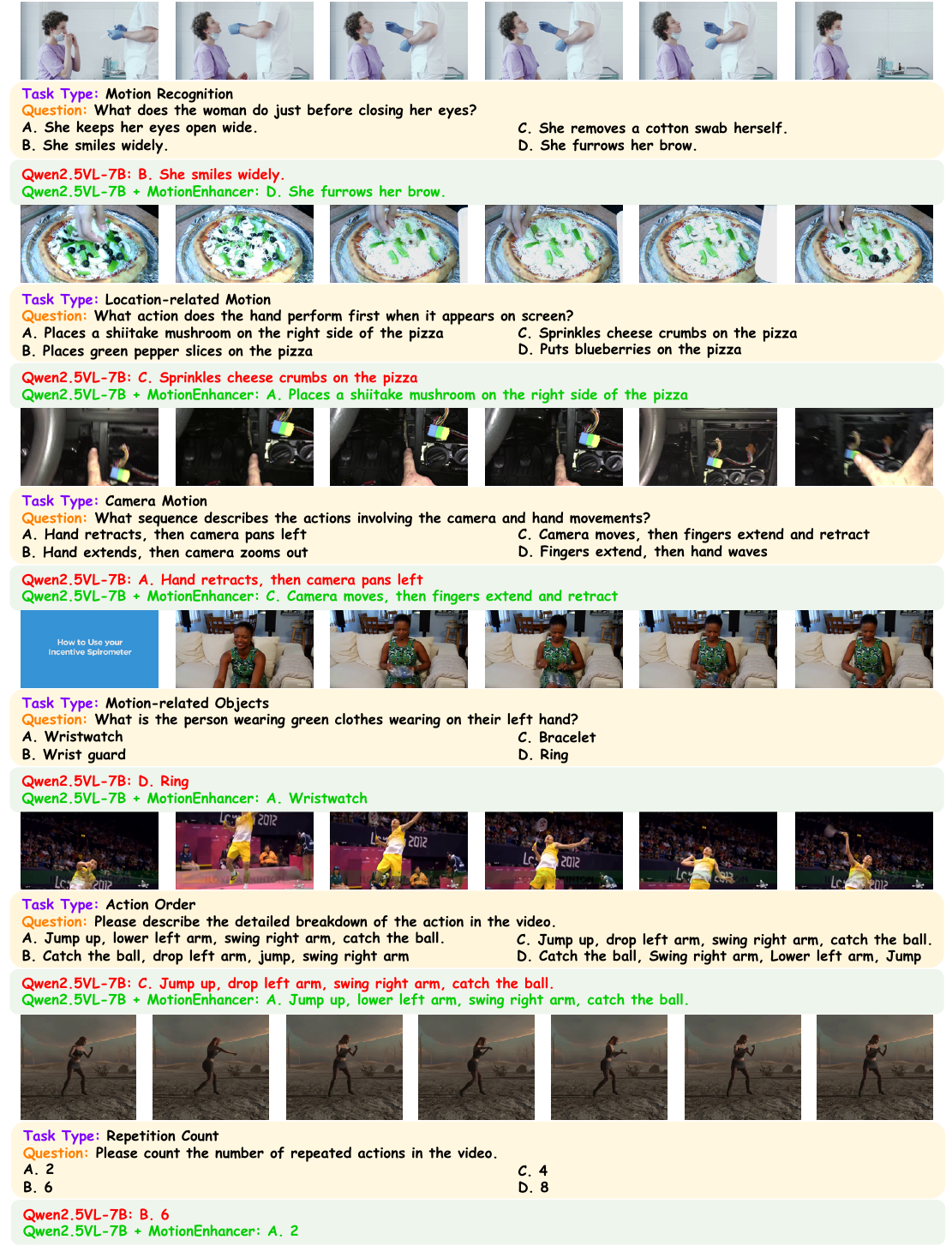}
    \caption{Qualitative results on MotionBench.}
    \label{fig:motionbench}
\end{figure*}

\begin{figure*}
    \centering
    \includegraphics[width=0.96\linewidth]{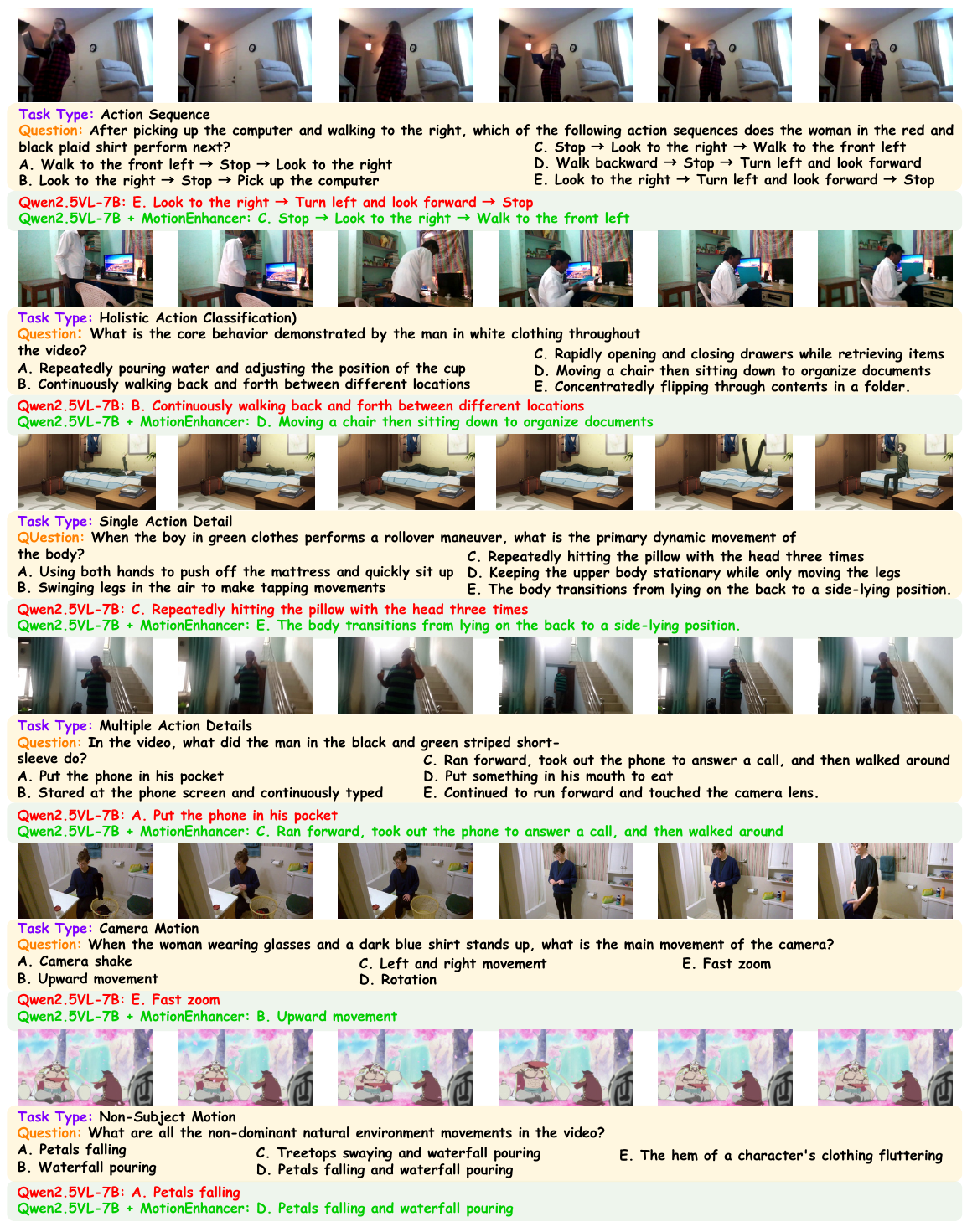}
    \caption{Qualitative results on FAVOR-Bench.}
    \label{fig:favorbench}
\end{figure*}

We present qualitative comparisons on MotionBench (Fig.~\ref{fig:motionbench}) and FAVOR-Bench (Fig.~\ref{fig:favorbench}) between the baseline Qwen2.5-VL-7B model and its \method{}-enhanced version across a range of motion-centric tasks.
These case studies underscore two key contributions of \method{}:
\textit{Improved temporal grounding:} the model better aligns textual descriptions with the actual sequence of events.
\textit{Enhanced motion sensitivity:} the model becomes more responsive to both foreground and background motion cues.
Importantly, these improvements are achieved without modifying the VLM architecture, highlighting the flexibility and generalizability of our attention alignment approach.

{\small
\putbib[main]
}

\end{bibunit}

\end{document}